\pdfoutput=1
\documentclass{article}
\usepackage{amsmath,graphicx,mlspconf}
\usepackage{xcolor}

\usepackage{amsfonts,amssymb,bm}
\usepackage[breaklinks=true]{hyperref}
\usepackage{amsthm}

\usepackage[ruled]{algorithm2e}
\newcommand{\quotes}[1]{``#1''}
\usepackage[short,nocomma]{optidef}
\usepackage[normalem]{ulem}
\theoremstyle{definition} 

\usepackage{booktabs}

\DeclareMathOperator*{\argmax}{arg\,max}
\DeclareMathOperator*{\argmin}{arg\,min}
\DeclarePairedDelimiter\ceil{\lceil}{\rceil}
\DeclarePairedDelimiter\floor{\lfloor}{\rfloor}

\title{QSMP: finding representative time series subsequences through Quick Shift+Matrix 
Profile}
\name{%
    Carlos H. Mendoza-Cardenas$^{\star\ddagger}$%
    \qquad Rogers F. Silva$^{\dagger}$%
    \qquad Austin J. Brockmeier$^{\ddagger\,\text{\S}}$%
}
\address{%
    \parbox{0.92\textwidth}{\centering
        $^{\star}$ Twitch Interactive, San Francisco, California, USA \\%
        $^{\dagger}$ TReNDS Center, Atlanta, GA, USA \\%
        $^{\ddagger}$ Department of Electrical and Computer Engineering,
        $^{\text{\S}}$ Department of Computer and Information Sciences,
        University of Delaware, Newark, DE, USA%
    }%
}

\newcommand{\arxivcopyright}{%
    \begingroup\renewcommand{\thefootnote}{}%
    \footnotetext{Accepted for publication in the 2026 IEEE International
    Workshop on Machine Learning for Signal Processing (MLSP).
    \copyright~2026 IEEE. Personal use of this material is permitted.
    Permission from IEEE must be obtained for all other uses, in any current or
    future media, including reprinting/republishing this material for
    advertising or promotional purposes, creating new collective works, for
    resale or redistribution to servers or lists, or reuse of any copyrighted
    component of this work in other works.}%
    \endgroup}

\begin{document}
\ninept
\providecommand{\arxivcopyright}{}

\maketitle
\arxivcopyright

\begin{abstract}
Finding representative waveforms in long time series has scientific and practical value 
in many domains, as it enables summarization and visualization of large time series datasets, 
and downstream tasks like classification and forecasting. We present here QSMP, a method to find 
representative waveforms in long time series through a density-guided clustering of time series 
subsequences. Our method makes a novel connection between Quick Shift, a mode-seeking algorithm, 
and the Matrix Profile, a time series similarity-search data structure, to adapt Quick Shift to 
the clustering of subsequences in long time series, with a space complexity that is superior to the state-of-the-art method. Our experiments on synthetic and real datasets show that
QSMP can be a valuable tool to summarize and visualize long time series by finding 
representative waveforms. 
\end{abstract}

\newcommand{\cem}[1]{\textcolor{blue}{cem: #1}}

\section{Introduction}

There are multiple domains where the shapes of representative waveforms in a time series
are correlated with a variable of interest, like the state of a brain disease from 
electroencephalographic (EEG) recordings \cite{Cole2017,isabel2026interpretable}, the feeding patterns of 
insects from electrical penetration graphs (EPG) \cite{Willett2016}, and the source
of an earthquake from continuous ground motion data \cite{Bergen2019}. Here, a waveform (a subsequence in the time series) is said to be \textit{representative} if it can sufficiently approximate, after appropriate scaling, a non-negligible fraction of subsequences of the time series. Such waveforms could be used in time series summarization \cite{Sripada2003} and 
visualization \cite{Balasubramanian2013, Rodrigues2021}, or in prediction tasks like seizure 
detection \cite{Mendoza-Cardenas2021}. Ultimately, an approximation is only sufficient if subsequent decisions made on the time-series by a human expert or information based on statistics of the time-series remain valid with the approximation.  

Although a wealth of methods have been proposed to find re-occurring waveforms in time series through 
motif discovery \cite{Mueen2014, Zhu2018a, Aghbari2020}, they do not take into account how much of 
the time series is represented, or covered, by a candidate waveform. Some 
methods have tackled this problem by clustering time series subsequences through the estimation of the 
density of each subsequence \cite{Denton2004, Minnen2007}, but their computational complexity makes them 
unfeasible for long time series. Snippet-Finder \cite{Imani2020} was 
proposed for finding representative waveforms (snippets), while also providing an estimate of the 
coverage of each snippet in the time series and the points in time where those snippets occur.

Here, we propose to find representative waveforms as the modes of the probability density estimate formed
from all subsequences, following an approach similar to \cite{Denton2004, Minnen2007}, but with superior 
scalability for long time series. To form density estimates efficiently for long time series, we propose 
a novel algorithm called QSMP (Quick Shift + Matrix Profile). Quick Shift is an efficient density-based 
clustering algorithm \cite{Vedaldi2008}, which exploits a hierarchical tree structure in order to scale 
the classic mean shift algorithm for density based clustering \cite{Fukunaga1975}, but typically requires 
storage of $\mathcal{O}(N^2)$, which is prohibitive for large $N$. The Matrix Profile is a framework for 
efficiently finding the nearest neighbors of all subsequences in long time series \cite{Yeh2016, Zhu2018, Zhu2018a}. 
Through careful design, our QSMP algorithm has a time and space complexity of $\mathcal{O}(N^2(B+3))$ and 
$\mathcal{O}(N)$, respectively, where $N$ is the number of subsequences in the time series and $B\ll N$ is the length of the min-pooling filter that we use to incorporate shift invariance into QSMP. QSMP is amenable to multi-GPU acceleration

Essentially, QSMP is a hierarchical, density-based clustering algorithm that performs shift-invariant clustering of time series subsequences, 
exploiting computational techniques from the Scalable Time series Ordered-search 
Matrix Profile (STOMP) algorithm \cite{Zhu2018} to perform Quick Shift 
efficiently in long time series.

QSMP operates by representing each subsequence of length $m$ in the time series (a window) as a point 
in $\mathbb{R}^m$. Subsequences that are similar will be close to each other in that 
$m$-dimensional space. Representative subsequences will have more neighbors within a small distance (i.e., the volume defined by a hypersphere); thus, representative subsequences will be found in dense regions. To find the set of subsequences in the densest regions, we first estimate the distribution of the windows in that $m$-dimensional space using kernel density estimation (KDE) 
\cite{Parzen1962}. Second, for each subsequence, we find the nearest neighbor (NN) 
whose density is higher, and save distance and index of that neighbor. We call 
that distance and index the NN-distance and NN-index, respectively. This method 
creates a directed acyclic graph (an anti-arborescence \cite{Gallier2011}), where each node is a subsequence in the time 	
series that points to its nearest neighbor. This graph can then be cut at edges 
where the NN-distance is above a given threshold, thus forming clusters, and 
the node with the highest density in each cluster is the \quotes{mode}, which we deem the most representative waveform of that cluster. After computing the density, NN-distance, and NN-index vectors, we can explore 
the different sets of modes corresponding to different thresholds, and examine 
the waveforms corresponding to each mode. All of this can be done \textit{post hoc}, as
there is no need to recompute those three vectors. The distance threshold can be set from the statistics of the data (e.g., the top quartile of the NN-distances gives a range of reasonable thresholds), so the number of clusters need not be specified but emerges from the more interpretable, post-hoc choice of threshold.


A key contribution is efficient scaling. A standard hierarchical clustering algorithm would require all pairwise distances, which is unfeasible for long time series, and many are trivial matches corresponding to overlapping windows starting at neighboring time points. 
A second contribution is that QSMP is an \textit{exact} method: all windows are analyzed. In contrast, \textit{approximate} algorithms use non-overlapping subsequences as candidates, so some representative subsequences may be split across two contiguous windows and missed. Code is available: \url{https://github.com/cniel-ud/qsmp}.

\section{Related work}
Some density-based methods have been proposed to find meaningful patterns in 
time series.  An early work \cite{Denton2004} proposed  
using a mode-seeking algorithm to cluster time series subsequences. There,
weighted kernel density estimation of the data is used, followed by the mean-shift 
algorithm to find meaningful patterns. In particular, a 
discrete random walk model of the time series is used to generate candidate 
random walk noise subsequences. The data is then weighted based on 
their similarity to those noise signals, aiming to de-emphasize the importance of 
noise-like subsequences in the estimation of the data density. The complexity of 
such method limits its application to small time series and short subsequences, 
due to the complexity of mean-shift and the exponential growth of the number 
of candidate random walk signals with the subsequence length. Other work \cite{Minnen2007} uses $k$-nearest neighbors search, a density estimate based on the distance to the $k$th nearest neighbor, and hidden Markov models for the local modes; their evaluation is again limited to small time series ($<$1 million samples). 

Another work closely related to QSMP is the Snippet-Finder algorithm 
\cite{Imani2020}. Snippet-Finder uses the Matrix Profile distance (MPdist) \cite{Gharghabi2020}, which compares two time series in terms of the similarity of their subsequences, 
to build a distance profile for the original time series to each non-overlapping subsequence in the time series.
It then greedily builds a list of $k$ representative patterns (snippets) such that 
the curve resulting from taking the minimum of their profiles has minimum area. 
In addition to the sub-subsequence length, denoted here as $S$, the user must specify the number of snippets, $k$, and their length $m$, where $m\ll N$. When $S = m$,
MPdist becomes the standard Euclidean distance. Snippet-Finder computes an MPdist profile with $\mathcal{O}(N)$ entries for each of the $N/m$ candidate windows, giving time $\mathcal{O}(N^2 S/m)$ and space $\mathcal{O}(N^2/m)$; the quadratic space makes it infeasible for long series (we could not run it on our multi-million-sample ECoG data), whereas QSMP needs only $\mathcal{O}(N)$ space.

Finally, finding the $k$ most representative waveforms could be approximated by clustering subsequences with shift-aware generalizations of $k$-means, like shift-invariant $k$-means (sikmeans)~\cite{Mendoza-Cardenas2021,isabel2026interpretable} or $k$-shape~\cite{Paparrizos2016}. However, their centroids are by definition averages and may not accurately reflect actual time series subsequences. 


\section{Methods}
To describe the proposed time series subsequence clustering algorithm, we first discuss mode-seeking clustering that rely on the pairwise kernel evaluations between subsequences and then discuss the efficient computation of the required distances for long time series. 

\subsection{Kernel density estimation}
Density-based clustering describes clusters of data 
points around modes of the multivariate density  from which the points are sampled. We want to find or seek those modes, but we first need to estimate the density from the set of data points $\mathcal{X} = \{\mathbf{x}_1, \ldots, 
\mathbf{x}_N\}\subset \mathbb{R}^m$. For the purposes of clustering, the density estimate does not need to be properly normalized. Given a dissimilarity function $d: \mathbb{R}^m \times \mathbb{R}^m \rightarrow \mathbb{R}_{+}$, we adopt a Gaussian-like kernel to form the density estimate $\hat{f}: \mathbb{R}^m \rightarrow \mathbb{R}_{+}$ at an arbitrary point $\mathbf{x}_k \in \mathbb{R}^m$ as 
\begin{equation}
\hat{f}(\mathbf{x}_k) = \sum_{j=1}^{N} \exp\left(-\frac{d(\mathbf{x}_k, \mathbf{x}_j)^2}{2\sigma^2} \right),
\label{eq:density_estimate}
\end{equation}
where $\sigma$ controls the kernel width. Evaluating this estimate at all $N$ sample points in $\mathcal{X}$ requires $\mathcal{O}(N^2)$ pairwise dissimilarity computations, which we denote as $D_{i,j}=d(\mathbf{x}_i, \mathbf{x}_j),\; i,j\in \{1,\ldots,N\}$.

\subsection{Mode-seeking clustering}
Given a density estimate, \textit{Mean Shift} clustering \cite{Fukunaga1975} iteratively shifts each data point by a vector proportional to the gradient of the density estimate until  each point reaches a local maximum. \textit{Medoid Shift} \cite{Sheikh2007} also moves each point in the direction of the gradient ascent, but avoids unsampled regions by restricting the 
shift vector to point towards a point that belongs to $\mathcal{X}$. In contrast, \textit{Quick Shift} \cite{Vedaldi2008} is a 
mode-seeking algorithm that clusters the data in a single iteration through the 
sample, by connecting each data point $\mathbf{x}_i$ to the nearest neighbor that has 
higher density, $\mathbf{x}_{v_i}$. The index $v_i$ is computed as
\begin{equation}
    v_i =\argmin_{{ j :\  \hat{f}(\mathbf{x}_j) > \hat{f}(\mathbf{x}_i)}}
	{d(\mathbf{x}_i,\mathbf{x}_j)},\quad i\in\{1,\ldots,N\}\setminus \{i^*\},
	{\label{eq:quick_shift_argmin}}
\end{equation}
with $i^*=\argmax_{i \in \{1,\ldots,N\}} \hat{f}(\mathbf{x}_i)$ being the point with the highest density of all. For completeness, we set $v_{i^*}=i^*$ to indicate the self-loop at the root. (We assume no ties in the dissimilarity or density, but note that both can be broken arbitrarily to ensure a fully connected tree.) We call the vector $\mathbf{v} = [v_1, \ldots, v_N]$ the NN-index vector, as it contains the indices of the nearest neighbor (NN) for each point (and a self-loop at the root), and the vector $\mathbf{d} =[d(\mathbf{x}_1, \mathbf{x}_{v_1}),\ldots, d(\mathbf{x}_N, \mathbf{x}_{v_N})] = [D_{1,v_1},  \ldots, D_{N,v_N}]$ the NN-distance vector, as its elements are the distance between each point and its nearest neighbor. Finally, with $\hat{\mathbf{f}} = [\hat{f}_1,\ldots,\hat{f}_N]= [\hat{f}(\mathbf{x}_1),  \ldots, \hat{f}(\mathbf{x}_N)]$, the output of the Quick Shift algorithm is grouped into what we call the \textit{QS-tuple}: 
$(\hat{\mathbf{f}}, \mathbf{d}, \mathbf{v})$. 
Quick Shift has a time complexity of $\mathcal{O}(mN^2)$, which is the same order as computing the distance matrix, with a small constant \cite{Vedaldi2008}. 

Essentially, Quick Shift (QS) builds a tree as a directed acyclic graph, where the nodes are the data points. There is a directed edge from each point to the nearest neighbor that has higher density, and the length of the edge is equal to the distance between the two points. Paths lead from each data 
point to the point with highest density (the root of the tree). Mathematically, a weighted tree can be represented as a collection of edges $T=\{e_1,\ldots,e_{|T|}\} \subseteq \{1,\ldots,N\}\times\{1,\ldots, N\}$ and associated weight $W:\{1,\ldots, N\} \times \{1,\ldots, N\} \rightarrow \mathbb{R}$. Quick Shift creates a tree $T_\text{QS}=\{(i,v_i)\}_{  i\neq v_i}$ with $W(i,v_i)=d_i=\min_{j \ :\  \hat{f}_j> \hat{f}_i} d(\mathbf{x}_i,\mathbf{x}_j)$.   

To perform 
clustering, we can then cut the tree at edges that are longer than a given 
distance threshold, $\tau$. Define the new NN-index and NN-distance as $\mathbf{v}'$ and $\mathbf{d}'$, by setting $v'_i = 1$ and $d_i'=0$ if $d_i > \tau$, converting the $i$th subsequence into a new root,  and $v_i'=v_i$ and $d_i'=d_i$ if $d_i\leq \tau$. Generally, this segments the original tree into a graph $G$ consisting of $k$ rooted trees, $G= \{(i,v_i)\}_{  i\neq v_i}=T_1 \cup T_2 \cup \cdots \cup T_k$. This segmentation effectively
clusters the data, with the root of each subtree being the representative waveform of each cluster. Smaller values of $\tau$ yield more trees with less edges, that is $\tau\leq \tau'\implies  G_\tau \subseteq G_{\tau ' } \subseteq T_\text{QS}$. 


In summary, $\sigma$ and $\tau$ play distinct roles: $\sigma$ sets the scale of the density estimate (different $\sigma$ yield different trees, and the useful scale can be application-specific), while $\tau$ controls how finely the tree is segmented. Because the z-normalized Euclidean distance~(\ref{eq:z-normalized_distance}) is unit-invariant, $\sigma\approx 1$ is a sensible starting point; given $\sigma$, we set $\tau$ by a binary search for the value yielding a desired number of modes. Absent an application-specific $\sigma$, we choose it without supervision: sweep a small grid, reusing the once-computed density and NN vectors, and keep the $\sigma$ whose modes are most mutually distinct (largest minimum pairwise distance).




\subsection{Time Series Subsequence Distance}
A times series, $\mathbf{x} = [x_1, x_2, \ldots, x_M] \in \mathbb{R}^M$, is a sequence of real values of length $M$, whereas a time series subsequence, $\mathbf{x}_i = [x_i, \ldots, x_{i+m-1}] \in \mathbb{R}^m$, is a sequence of real values of length $m < M$, starting at time step $i$.  Let $\mathcal{X} = \{\mathbf{x}_i\}_{i=1}^N$ be the set of all subsequences in $\mathbf{x}$, with $N = M-m+1$. To ameliorate low-frequency wanderings and amplitude variation, one can use the z-normalized Euclidean dissimilarity~\cite{keogh2002need} to compare two length-$m$ subsequences as
\begin{align}
d(\mathbf{x}_i, \mathbf{x}_j) = \left \lVert\frac{\mathbf{x}_i\!-\!\mu_i}{\sigma_i}-\frac{\mathbf{x}_j\!-\!\mu_j}{\sigma_j} \right \rVert_2,
\label{eq:z-normalized_distance}
\end{align}   
where  $\mu_i=\frac{1}{m}\sum_{n=0}^{m-1}x_{i+n}$ and $\sigma_i=\sqrt{\frac{1}{m}\sum_{n=0}^{m-1}(x_{i+n}-\mu_i)^2}$ are the mean and 
standard deviation of the values in $\mathbf{x}_i$, and likewise $\mu_j$ and 
$\sigma_j$ are the mean and standard deviation of $\mathbf{x}_j$, which can be computed efficiently for the entire time series. Let $\boldsymbol{\mu}, \boldsymbol{\sigma} = \texttt{rollingMeanAndStd}(\mathbf{x}, 
m)$ be the procedure that computes the mean, $\boldsymbol{\mu} \in 
\mathbb{R}^N$, and standard deviation, $\boldsymbol{\sigma} \in 
\mathbb{R}^N$, of all the $N$ subsequences in $\mathbf{x}$. Using the Welford's method 
\cite{Welford1962},  
\texttt{rollingMeanAndStd}   has a time complexity of $\mathcal{O}(N)$.

\subsection{Optimizing Quick Shift for Time Series Subsequences}
Our goal is to find a set of representative waveforms in $\mathbf{x}$ 
via Quick Shift density-based clustering of time series subsequences, such that many subsequences are within $\epsilon$-balls of the waveforms. To do so, we propose several modifications from standard Quick Shift.
Following the estimation of an empirical density function from the data points in $\mathcal{X}$ for $i \in \{1, \ldots, N\}$, we find the index of the nearest neighbor with higher density for each subspace $\mathbf{x}_i$ as:
%
%
\begin{argmini}
	{\substack{j \in \{ 1,\ldots,N: \tilde{f}_j > \tilde{f}_i  \}\setminus\{i-m/4,\ldots,i+m/4\} }}
	{\breve{D}_{i,j},}
	{\label{eq:quick_shift_argmin_recall_1}}{v_i =}
\end{argmini}
where the decorated functions indicate key modifications from standard Quick Shift. In total, there are five modifications to the computation of the nearest neighbor, density estimates, and Quick Shift trees, as follows.
First, when computing pairwise distances we ignore trivial matches: $d(\mathbf{x}_i,\mathbf{x}_j) := \infty$ if  $i - m/4 \le j \le i + m/4$, which is known as an \emph{exclusion zone} in previous literature \cite{Yeh2016}.

\begin{figure}[htb]
	\centering
	\includegraphics[width=0.65\linewidth]{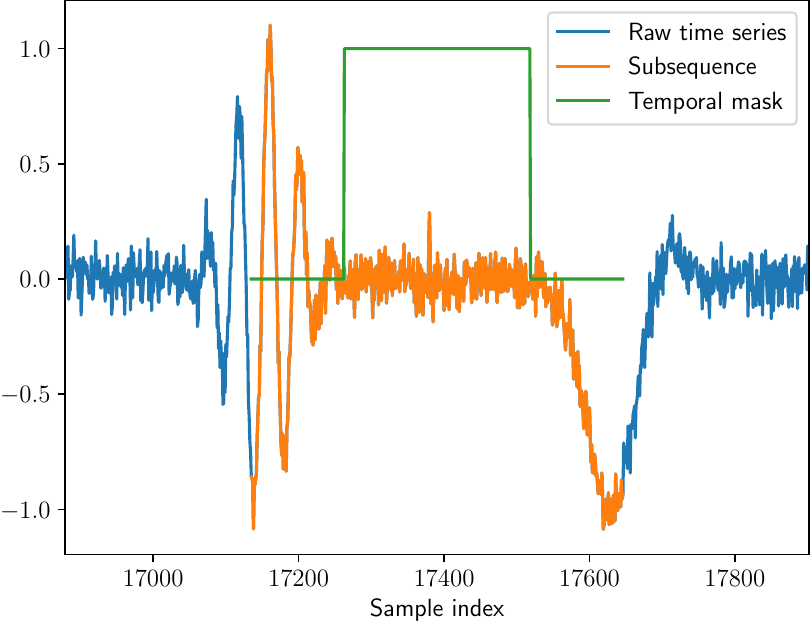}
	\caption[Shift-invariant distance]{Temporal mask (green) used to compute the centeredness factor. A well-centered window captures a full waveform, whereas the off-center orange window straddles two waveforms and captures only fragments; centeredness down-weights the latter.}
	\label{fig:centeredness_demo}
\end{figure}

Second, when the duration of the pattern of interest is smaller than the subsequence length, we want to ignore any subsequences where the pattern of interest occurs at the \emph{extremes} of the subsequence and away from its center. Fig.~\ref{fig:centeredness_demo} shows a toy example of two waveforms with short duration, and a subsequence (in  orange) that is capturing half of each waveform. We resolve this issue by locally changing the kernel width in the computation of the modified density $\tilde{f}(\mathbf{x}_i)= \sum_{j=1}^{N} \exp\left(-\frac{d(\mathbf{x}_i, \mathbf{x}_j)^2}{2(\sigma\kappa_i)^2} \right)$, with $\kappa_i = \tilde{\sigma}_i/\sigma_i$, $\tilde{\sigma}_i$ being the standard deviation of $\tilde{\mathbf{x}}_i = \mathbf{w} \odot \mathbf{x}_i$, and $\mathbf{w}$ being a rectangular window that is 1 at the center of the sequence and 0 elsewhere (green line in Fig.~\ref{fig:centeredness_demo}). We normalize $\boldsymbol{\kappa} = [\kappa_1, \ldots, \kappa_N]$ by its maximum value so that $\kappa_i \in (0, 1]$. We call $\kappa_j$ the \textit{centeredness} of $\mathbf{x}_i$. With this scaling of the distance, subsequences with small centeredness value will have small density value, effectively ignoring them in the next stages of Quick Shift so that off-center windows (like the orange one) are unlikely to become modes. This correction is needed because a window that captures only part of a waveform is dominated by a flat segment; after z-normalization such a window is rescaled and becomes spuriously close to many other low-amplitude windows, inflating its density.

Third, for the case of lacking continuity in the time series due to missing data, we discard the missing time points and concatenate the remaining segments into a single time series. Then we want to ignore any subsequences that are close to the locations of the splicing of two segments. If $\mathbf{s} = [s_1, s_2,  \ldots]$ is the vector of splice indices (one per missing-data segment), we set $\tilde{f}(\mathbf{x}_i) = 0$ if $s_i - m + 1 \le i \le s_i - 1$, for $i \in [M]$, which prevents neighboring segments from being selected by Quick Shift.  

\begin{figure}[htb]
	\centering
	\includegraphics[width=0.567435\linewidth]{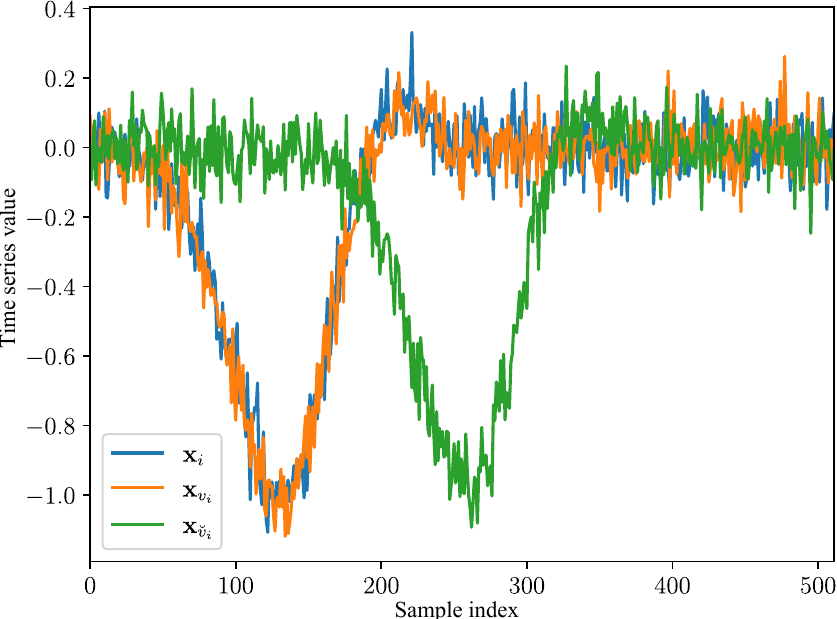}\\
~    \\
	\includegraphics[width=0.56745\linewidth]{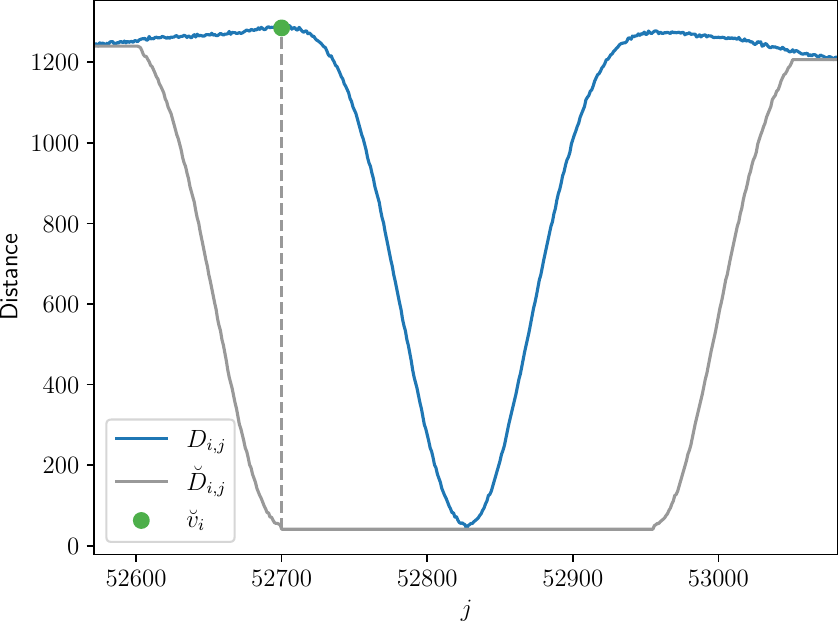}
	\caption[Shift-invariant distance]{(Top) The 
		reference subsequence $\mathbf{x}_i$ would be matched with its nearest 
		neighbor with higher density ($\mathbf{x}_{v_i}$), but there is a 
		subsequence that is closer 	($\mathbf{x}_{\breve{v}_i}$) if we 
		introduce 
		shift-invariance. (Bottom) The regular z-normalized Euclidean distance 
		($D_{i,j}$), and its shift-invariant version ($\breve{D}_{i,j}$), in 
		the 
		vicinity of $\breve{v}_i$.}
	\label{fig:QSMP_shift-invariance_demo}
\end{figure}

Fourth, we want to identify subsequences that could be aligned through a temporal shift/lag, but the distance between two subsequences does not account for partial matching within the sequence. Fig.~\ref{fig:QSMP_shift-invariance_demo} 
illustrate this situation. The blue line is 
distance between $\mathbf{x}_i$ (blue waveform) and $\mathbf{x}_j$, for a subset of indices $j$. Let $j^*$ be the local 
minimum of $D_{i,j}$ for a given $i$. At an optimal lag, $\mathbf{x}_i$ and 
$\mathbf{x}_{j^*}$ are well aligned and their distance is close to zero, and 
as we move away from $j^*$, $D_{i,j}$ grows rapidly.
We thus adopt a dissimilarity that allows local temporal translations of a time series subsequence 
\begin{equation}
\breve{D}_{i,j} = \min([D_{i,j-\floor*{B/2}}, 
D_{i,j-\floor*{B/2}+1}, \ldots, D_{i,j+\ceil*{B/2}-1}]),    
\end{equation}
and $B$ being the 
length of a min-pooling filter. 
Following the example in 
Fig.~\ref{fig:QSMP_shift-invariance_demo}, we can see the effect of 
using $\breve{D}_{i,j}$ instead of $D_{i,j}$ in the nearest-neighbor calculation in Quick Shift.
After min-pooling, the distance between subsequences $\mathbf{x}_i$ (blue waveform) and 
$\mathbf{x}_{\breve{v}_i}$ (green waveform) is drastically reduced with the latter becoming tied as the nearest neighbor of the former, demonstrating the shift invariance. 

Fifth, and last, after the tree is cut by picking a distance threshold $\tau$, neighboring subsequences may appear as distinct roots due to the exclusion zone. Thus, we further post-process a cut tree by merging subtrees whose roots are within the exclusion zone.   


\subsection{Efficient Computation of Quick Shift for Subsequences}

For a long time series $N$ is large, often in the millions, and the naive computation of the QS-tuple appears prohibitive. We build on time series data mining approaches \cite{Yeh2016, Zhu2018, Zhu2018a} to efficiently compute the distance matrix $\mathbf{D}=[D_{i,j}]_{i,j=1}^{N,N}$ and the NN-index $\mathbf{v}$, known as the Matrix Profile (MP) and MP index. Specifically, we adapt the STOMP algorithm \cite{Zhu2018} to compute the QS-tuple $(\tilde{\mathbf{f}}, \mathbf{d}, \mathbf{v})$ with time and space complexity $\mathcal{O}(N^2(B+3))$ and $\mathcal{O}(N)$, where $B<m$ is the min-pooling filter length used for shift invariance, and we exploit the problem's parallelism for GPU acceleration.

The naive computation of the density estimate (\ref{eq:density_estimate}) with the Euclidean distance has time and space complexity $\mathcal{O}(mN^2)$ and $\mathcal{O}(N)$. We exploit core ideas from STOMP to reduce the time complexity to $\mathcal{O}(N^2)$ and parallelize it on GPUs.

First we note that the squared z-normalized Euclidean dissimilarity can be expressed as $d^2(\mathbf{x}_i,\mathbf{x}_j)= {
	2m\left(1-\frac{\rho_{i,j} - m\mu_i\mu_j}{m\sigma_i\sigma_j}\right)}$, where $\rho_{i,j} = \mathbf{x}_i^\text{T}\mathbf{x}_j$ is the inner 
product between the subsequences. The key observation behind the STOMP algorithm is that the dot 
product $\rho_{i,j}$ can also be computed from
$\rho_{i-1,j-1}$ in $\mathcal{O}(1)$ time, since 
\begin{equation}
\rho_{i,j} = \rho_{i-1,j-1} + x_{i+m-1}x_{j+m-1}-x_{i-1}x_{j-1}.
\label{eq:recursive-dot-product-compute}
\end{equation} 
Consequently, $\boldsymbol{\rho}_2$, where $\boldsymbol{\rho}_i = [\rho_{i,1}, \rho_{i,2}, \ldots, \rho_{i,N}]$, can be computed from $\boldsymbol{\rho}_1$ in $\mathcal{O}(N)$, $\boldsymbol{\rho}_3$ from $\boldsymbol{\rho}_2$ in $\mathcal{O}(N)$, and so on. (The initial $\boldsymbol{\rho}_1$ can be computed with $\mathcal{O}(Nm)$ time and space complexity using matrix-vector multiplication or $\mathcal{O}(N\log(N))$ complexity as a convolution, through the fast Fourier transform (FFT) \cite{Yeh2016}. Note that 
$\rho_{i,1} = \rho_{1,i}$, so the first entry of $\boldsymbol{\rho}_i$ is just 
the $i$th entry of $\boldsymbol{\rho}_1$.) Each of the  $\mathcal{O}(N)$ is embarrassingly parallel. To see this we use two variables, $\boldsymbol{\rho}_\text{in}$ and 
$\boldsymbol{\rho}_\text{out}$, to store $\boldsymbol{\rho}_{i-1}$ and 
$\boldsymbol{\rho}_{i}$, respectively. 
GPUs usually have hundreds to 
thousands of cores, which allows us to split the computation of 
$\boldsymbol{\rho_\text{out}}$ into multiple concurrent threads, each thread 
reading from a small set of cells in $\boldsymbol{\rho_\text{in}}$ and writing 
to its corresponding set of cells in $\boldsymbol{\rho_\text{out}}$. With $P$ cores the time complexity is $\mathcal{O}(N (N/P))$.  
Likewise, the density computation \eqref{eq:density_estimate} can be performed across the same iterations. 
We implemented our algorithms by adapting the GPU-accelerated STOMP in the Python library STUMPY \cite{Law2019}; the full procedures are given in the supplementary material. The min-pooling filter costs $\mathcal{O}(BN)$ and is also GPU-parallelizable, so the final complexity of computing the NN-index and NN-distance is $\mathcal{O}(N^2(B+2))$.
We thus have that the total time and space complexity required to compute the 
QS-tuple is $\mathcal{O}(N^2(B+3))$ and $\mathcal{O}(N)$, respectively. 
Importantly, note that we can trivially compute, in parallel, the QS-tuple for multiple 
values of the kernel width, $\sigma$. That is, we can compute 
the density estimate for $n_\sigma$ different values of $\sigma$ as long as $n_\sigma \ll N$. 
The time complexity goes from $\mathcal{O}(N^2(B+2)$ to 
$\mathcal{O}(N^2(B+1+n_\sigma)$ and the space complexity increases to $\mathcal{O}(n_\sigma N)$.

\section{Experiments and Results}

We now present results on synthetic and real data to illustrate different characteristics of QSMP. Taking inspiration from EEG data, which has a 1/f spectrum (i.e., it falls off inversely proportional to 
the signal frequency), we tested the performance of QSMP on a synthetic 
time series with similar power-law distribution of frequencies. We 
generated six 1-second Morlet wavelets at frequencies $[1, 5, 12, 30, 100, 
150]$ Hz, sampled at 512 Hz. Then, we built a time series that is 
1,000 seconds long by concatenating 1,000 wavelets drawn at random from a 
power-law distribution over the six frequencies, and adding Gaussian noise with 
zero mean and standard deviation of 0.07. 
We refer to this time series as the \texttt{power-law} dataset.




For real data, we used the dataset Study019, a multi-channel epileptic ECoG recording that is publicly available
in the  \url{https://www.ieeg.org/} platform. This dataset is sampled at 512 Hz and preprocessed 
using the same pipeline introduced in \cite{Mendoza-Cardenas2021}. The recording is split into 
two segment classes: 1) 1-hour segments that are just before one of the 16 seizures present in the recording, 
called \emph{preictal} data, and 2) segments that are 4 hours away from any seizure, called \emph{interictal} data. 
Furthermore, the spatial channels of the multi-channel ECoG signal are linearly combined to maximize 
the energy of one class (preictal/interictal), while minimizing the energy of the other class, effectively
transforming the multi-channel signal into a single-channel time series. We took here the first 
$n=5,794,755$ and $n=5,644,856$ samples of preictal and interictal classes, respectively, corresponding 
to more than 3 hours of ECoG recording per class.



\subsection{Results on Synthetic Data}
All methods use $m=512$ and target $k=6$ patterns, with each method's remaining
hyperparameter chosen \emph{without supervision} by its own objective (no ground
truth used): QSMP keeps the $\sigma\in\{0.5,0.9,1,2,3\}$ whose $k$ modes are most
mutually distinct ($B=256$; $\tau$ binary-searched for six modes); sikmeans keeps
the smallest-distortion window length $L\in\{640,768,1024\}$ (30 restarts each);
Snippet-Finder keeps the best-covering sub-subsequence length $S\in\{77,\ldots,256\}$
($15$--$50\%$ of $m$).
%
We quantify recovery over 20 random realizations of the dataset, scoring how
well each method recovers the six known prototypes. Each returned pattern is matched (with replacement) to its closest
ground-truth prototype using a shift-invariant distance, and we report three
metrics (mean~$\pm$~95\% CI): the number of distinct frequencies recovered
(FreqRec, of 6); the mean shift-invariant cosine similarity of the matched
patterns (CosSim, $1$ = identical morphology); and the mean peak-frequency
error (PeakErr). Table~\ref{tab:recovery} shows QSMP recovers nearly all
frequencies with the highest morphological fidelity, and is best on every
metric ($p<10^{-3}$, paired $t$-test). As a fairness check, scoring recovery
under Snippet-Finder's own MPdist~\cite{Gharghabi2020} yields the same ordering
(QSMP $1.13$ vs.\ $2.59$ vs.\ $3.48$; $p<10^{-3}$), so the gap is not an
artifact of our distance choice.

\begin{table}[htb]
    \centering
    \caption{Ground-truth recovery on 6 waveforms on the \texttt{power-law} dataset (mean
    $\pm$ 95\% CI over 20 seeds). $\uparrow$/$\downarrow$: higher/lower is
    better.}
    \label{tab:recovery}
    \resizebox{\linewidth}{!}{%
    \begin{tabular}{lccc}
    \toprule
    Method & FreqRec $\uparrow$ & CosSim $\uparrow$ & PeakErr (Hz) $\downarrow$ \\
    \midrule
    QSMP & \textbf{5.9$\pm$0.1} & \textbf{0.97$\pm$0.02} & \textbf{1.3$\pm$2.6} \\
    Snippet-Finder & 3.6$\pm$0.4 & 0.87$\pm$0.04 & 33.1$\pm$6.7 \\
    sikmeans & 2.9$\pm$0.2 & 0.76$\pm$0.05 & 43.2$\pm$4.3 \\
    \bottomrule
    \end{tabular}}
\end{table}

\begin{figure}[htb]
    \centering
    \includegraphics[width=1.0\linewidth]{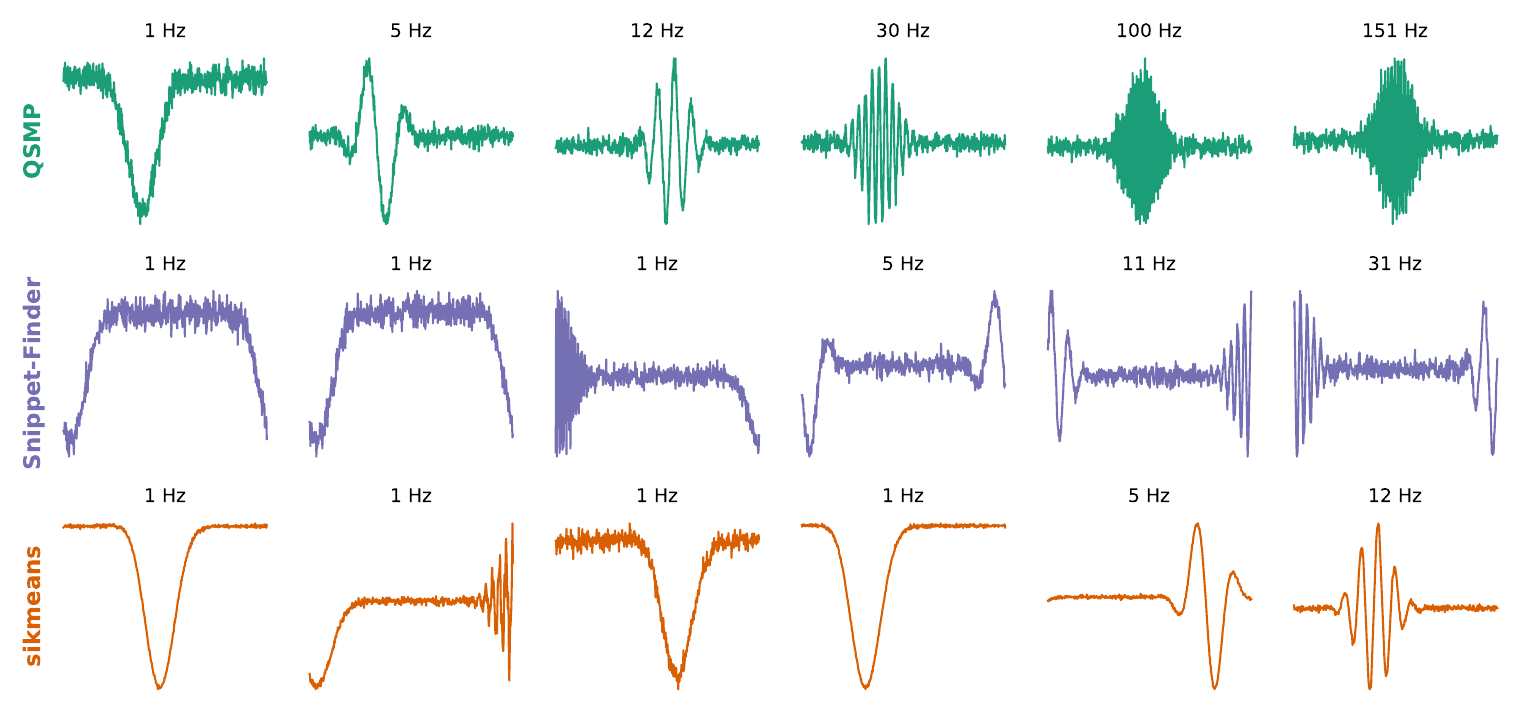}
    \caption{Patterns returned by each method on a representative realization of the
    \texttt{power-law} dataset (unsupervised selection), labeled by estimated peak
    frequency. QSMP recovers all six frequencies cleanly; Snippet-Finder and sikmeans
    duplicate the prevalent low frequencies and miss the rare high ones.}
    \label{fig:QSMP_vs_SnippetFinder_vs_sikmeans_Morlet}
\end{figure}

Fig.~\ref{fig:QSMP_vs_SnippetFinder_vs_sikmeans_Morlet} illustrates these results on a
representative realization.

We further stress-test the methods on a harder variant in which wavelet activation times are uniformly distributed as in a Poisson point process, with exponentially distributed intervals; those
extended results and analyses are in the supplementary material.

\subsection{Results on ECoG data}
\begin{figure}[tbh]
	\centering
	\includegraphics[width=0.43\linewidth]{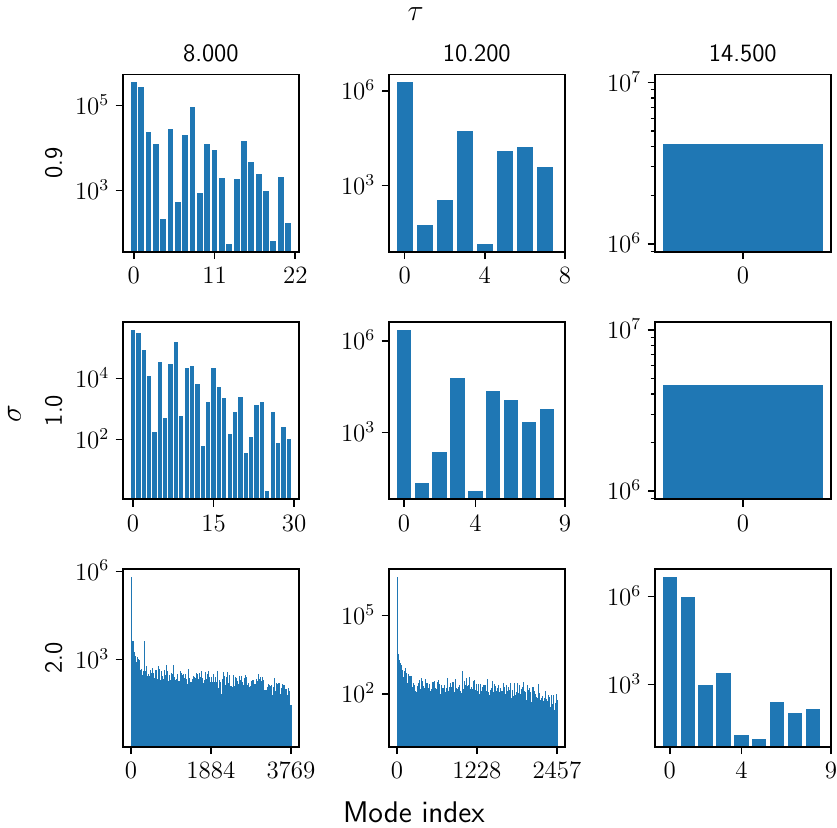}\quad
	\includegraphics[width=0.51\linewidth]{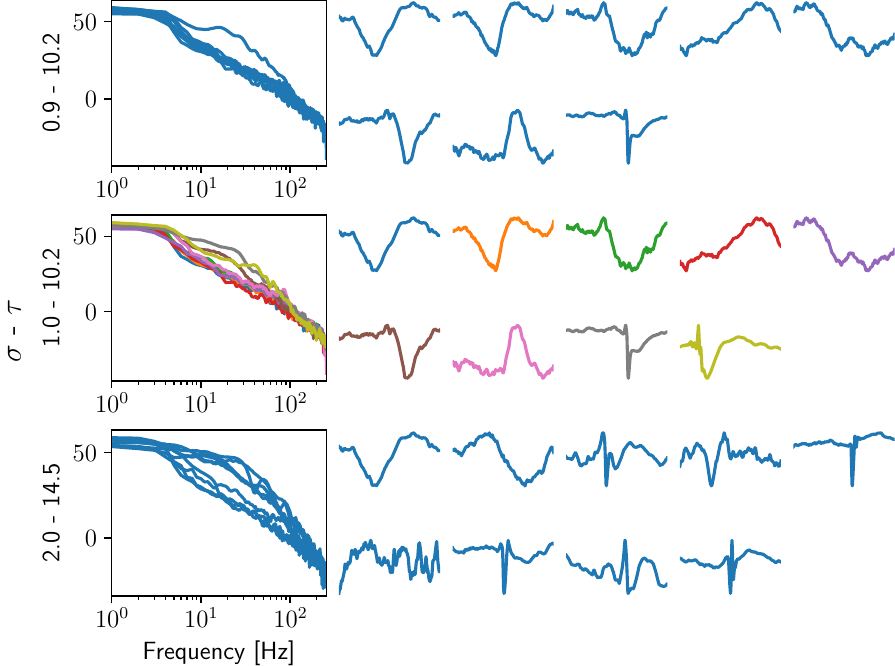}
	\caption[Effect of $\sigma$ and $\tau$ on Study019-preictal dataset]{Effect 
	of 
		$\sigma$ and $\tau$ on Study019-preictal dataset. 
		(left) Distribution of modes. (right) Mean PSD of each mode, averaged 
		across the five nearest neighbors, with the modes shown to the right of 
		each PSD. The PSD curves and modes in the middle row are matched by 
		color.}
	\label{fig:hists_PSDs_study019-preictal}	
\end{figure}

We now apply QSMP to the ECoG data, with a subsequence length of $m=512$ corresponding to 1 second. We believe that this is a sensible choice for EEG/ECoG data, as most of the relevant  neural activity occurs at frequencies that are above 1 Hz. The diversity of prototypical waveform shapes in ECoG data  is large, as seen in Fig.~\ref{fig:hists_PSDs_study019-preictal}. %
Even though ECoG activity is dominated by 
low-frequency waveforms due to the 1/f law of its PSD, QSMP is 
able to recover several high-frequency patterns. In the right pane, second row of 
Fig.~\ref{fig:hists_PSDs_study019-preictal}, for example, the gray and
olive green patterns are well-known to be related with seizure activity, known as
``sharp spikes'', and ``spikes and waves'', respectively~\cite{Westmoreland1996}. We stress that QSMP is a data-driven discovery tool: it surfaces recurrent morphologies for expert inspection, and confirming their clinical significance requires downstream validation beyond the present scope. Fig.~\ref{fig:hists_PSDs_study019-preictal} also shows how $\sigma$ and $\tau$ control the number and diversity of modes: since distances are unaffected by $\sigma$, a larger width lets more diverse (higher-frequency) waveforms gain density and survive as modes.

\begin{figure}
		\centering
		\includegraphics[width=0.48\linewidth]{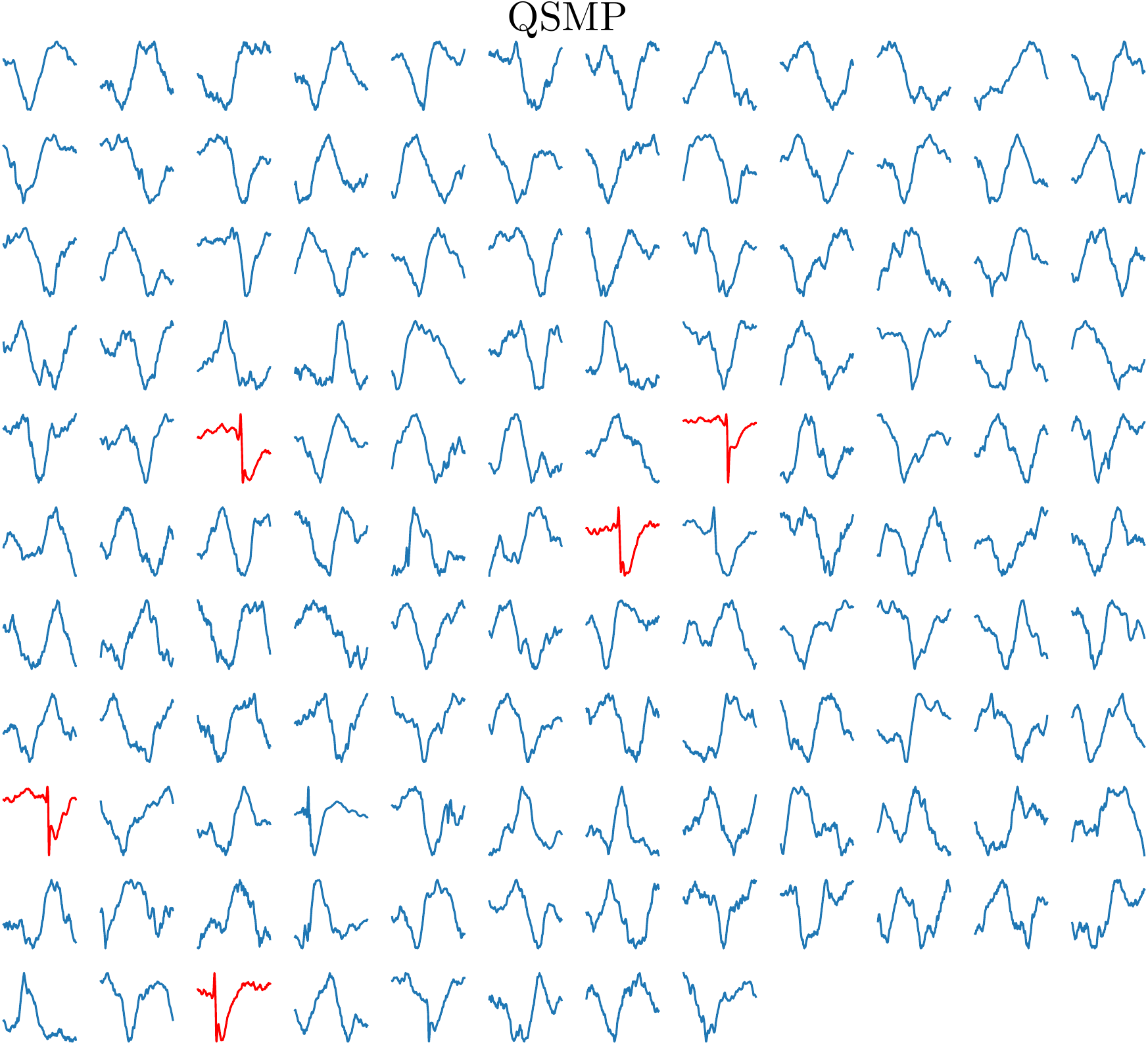}\quad
		\includegraphics[width=0.48\linewidth]{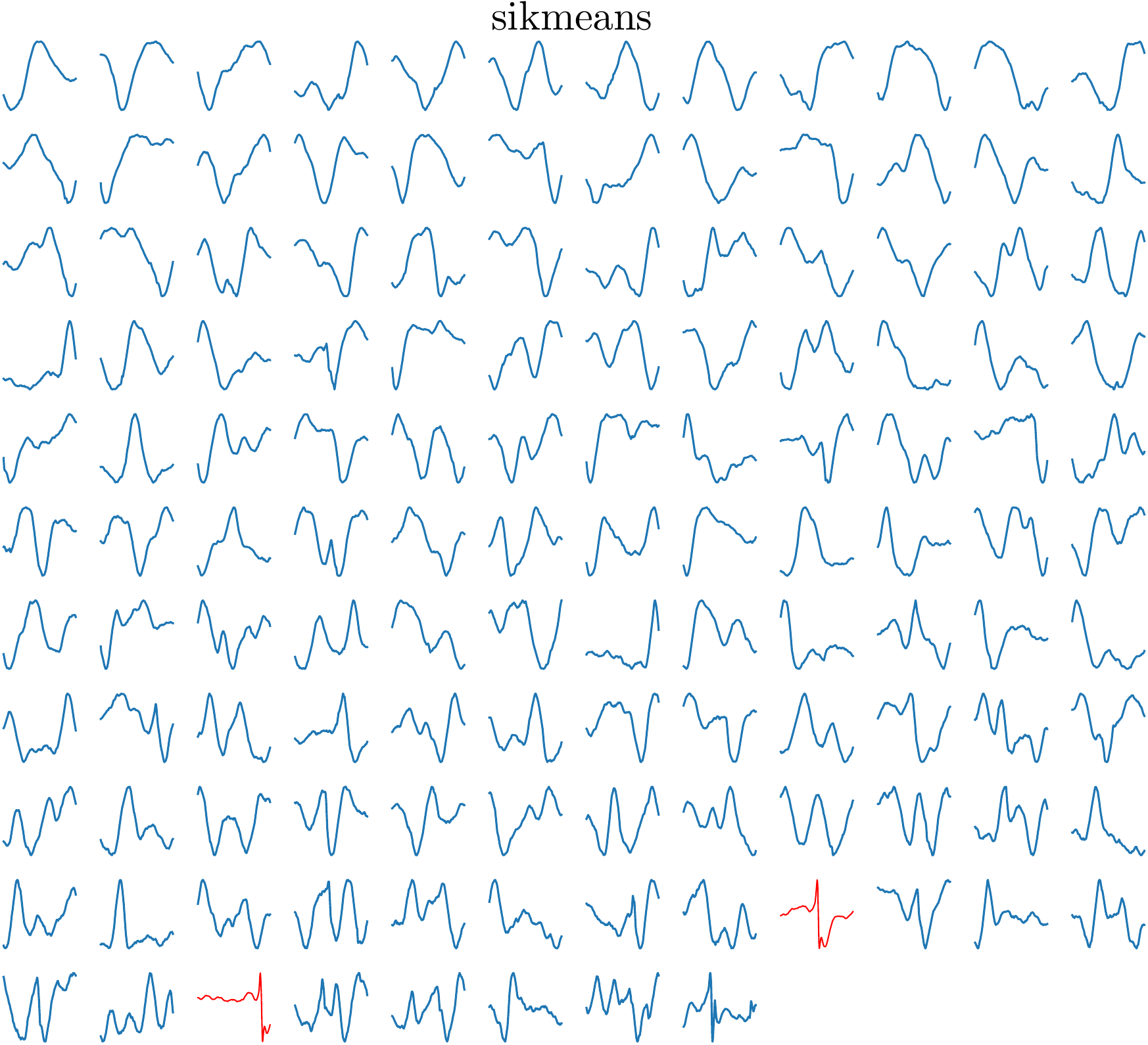}
		\caption[QSMP and sikmeans patterns in the Study019-preictal 
		dataset.]{QSMP and 
			sikmeans patterns in the Study019-\emph{preictal} dataset. We manually 
			highlight in 
			red high-frequency epileptiform patterns to emphasize that QSMP is 
			more 
			sensitive to changes in pattern morphology.}
		\label{fig:study019-preictal_waves}	
\end{figure}

We also compare QSMP and sikmeans qualitatively here (Snippet-Finder could not be run). Both use $m=512$ and $k=128$ patterns; QSMP uses $\sigma=1.0$ with $\tau$ set for 128 modes, and sikmeans a window length $L=768$. Fig.~\ref{fig:study019-preictal_waves} shows the \emph{preictal} patterns for both. As expected, the averaged patterns in sikmeans are smoother than the raw
patterns in QSMP. Note that averaging of a QSMP mode with its nearest neighbors can reduce noise if desired (discussed in the supplementary material). 
The \emph{interictal} Study019 data yields the same qualitative contrast---QSMP surfaces sharper, higher-frequency morphologies than sikmeans' smoothed centroids (shown in the supplementary material).

\subsection{Downstream evaluation on MixedBag}
To quantify motif quality on a downstream task, we use the MixedBag benchmark~\cite{Imani2020} (100 series, each split into two regimes at a known change point). We cut the QSMP tree to two modes and count a \emph{success} when they fall in different regimes, using the subsequence length $\check{m}_i$ provided with each series. To avoid tuning to the ground truth, each method's remaining hyperparameter is chosen by its own \emph{unsupervised} criterion, as in the recovery experiment: for QSMP the kernel width $\sigma$ maximizing the separation between the two modes, and for sikmeans the window length minimizing clustering distortion; for Snippet-Finder we cite the authors' result at their fixed setting ($S=50\%$)~\cite{Imani2020}. Table~\ref{tab:success_rate_comparison} shows QSMP is competitive with Snippet-Finder and far above sikmeans and random sampling, despite its much smaller space complexity. This benchmark rewards \emph{coverage}---one representative per regime---which Snippet-Finder is designed to optimize; QSMP matches it here while also leading on morphological fidelity (Table~\ref{tab:recovery}). As an uncontrolled timing reference, one MixedBag run took $\sim$15\,min for QSMP (single GPU) vs.\ $>$8\,h for the authors' CPU Snippet-Finder.
\begin{table}[htb]
    \centering
    \caption{Success rate (\%) on the MixedBag dataset (100 series;
    $\pm$ 95\% CI under a Bernoulli model, $z\sqrt{p(1-p)/100}$).
    Snippet-Finder as reported in~\cite{Imani2020}; QSMP and sikmeans use
    unsupervised hyperparameter selection; random sampling is the analytical
    baseline of~\cite{Imani2020}.}
    \begin{tabular}{cccc}
    \toprule
       QSMP &  Snippet-Finder & sikmeans & random\\
       \midrule
       82.0$\pm$7.5 & 84.0$\pm$7.2 & 38.0$\pm$9.5 & 47.3$\pm$9.8\\
       \bottomrule
    \end{tabular}
    \label{tab:success_rate_comparison}
\end{table}

\section{Conclusion}

We presented QSMP, a method that finds representative patterns in very long time series with linear space complexity, quadratic time complexity, and multi-GPU parallelization. Its modes are actual subsequences from the raw time series. For a domain expert, such as a neurologist examining EEG, this is highly desirable, enabling visualization of raw patterns and their temporal location, both available from QSMP. 
Furthermore, one can compute the QS-tuple for multiple values of the kernel width, $\sigma$, without
significantly increasing the computational complexity, which reduces the burden
of finding an optimal $\sigma$. Once the QS-tuple has been computed, one can
cut the QS tree \textit{post hoc} at different combinations of $\sigma$ and $\tau$; given $\sigma$, a binary search can automatically find the $\tau$ that yields a desired number of patterns. QSMP thus enables \textit{visual}, exploratory discovery of representative patterns in long time series, letting domain experts steer the final clustering selection.


\bibliographystyle{IEEEbib}
{\footnotesize
\bibliography{ThirdPaper.bib,strings}

@article{isabel2026interpretable,
  title={Interpretable EEG biomarkers for neurological disease models in mice using bag-of-waves classifiers},
  author={Isabel Cano Achuri, Maria and Kay Lara, Montana and Abed Rabbo, Khalil and Wilson, Benjamin T and Meek, Austin and Mahoney, J Matthew and Hernan, Amanda E and Brockmeier, Austin J},
  journal={Journal of Neural Engineering},
  volume={23},
  number={3},
  pages={036016},
  year={2026},
  publisher={IOP Publishing}
}

@inproceedings{keogh2002need,
  title={On the need for time series data mining benchmarks: a survey and empirical demonstration},
  author={Keogh, Eamonn and Kasetty, Shruti},
  booktitle={Proceedings of the Eighth ACM SIGKDD International Conference on Knowledge Discovery and Data Mining},
  pages={102--111},
  year={2002}
}

@article{Sripada2003,
author = {Sripada, Somayajulu G. and Reiter, Ehud and Hunter, Jim and Yu, Jin},
doi = {10.3115/1067737.1067775},
journal = {10th Conf. Eur. Chapter Assoc. Comput. Linguist. EACL 2003},
pages = {167--170},
title = {{Summarizing neonatal time series data}},
year = {2003}
}

@article{Denton2004,
author = {Denton, Anne},
journal = {3rd Int. Work. Min. Temporal Seq. Data},
pages = {14--21},
title = {{Density-based Clustering of Time Series Subsequences}},
year = {2004}
}

@article{Mueen2014,
author = {Mueen, Abdullah},
doi = {10.1002/widm.1119},
issn = {19424787},
journal = {Wiley Interdiscip. Rev. Data Min. Knowl. Discov.},
number = {2},
pages = {152--159},
title = {{Time series motif discovery: Dimensions and applications}},
volume = {4},
year = {2014}
}

@article{Yeh2016,
author = {Yeh, Chin Chia Michael and Zhu, Yan and Ulanova, Liudmila and Begum, Nurjahan and Ding, Yifei and Dau, Hoang Anh and Silva, Diego Furtado and Mueen, Abdullah and Keogh, Eamonn},
doi = {10.1109/ICDM.2016.89},
isbn = {9781509054725},
issn = {15504786},
journal = {Proc. - IEEE Int. Conf. Data Mining, ICDM},
pages = {1317--1322},
publisher = {IEEE},
title = {{Matrix profile I: All pairs similarity joins for time series: A unifying view that includes motifs, discords and shapelets}},
year = {2016}
}

@article{Willett2016,
author = {Willett, Denis S. and George, Justin and Willett, Nora S. and Stelinski, Lukasz L. and Lapointe, Stephen L.},
doi = {10.1371/journal.pcbi.1005158},
issn = {15537358},
journal = {PLoS Comput. Biol.},
number = {11},
pages = {1--14},
pmid = {27832081},
title = {{Machine Learning for Characterization of Insect Vector Feeding}},
volume = {12},
year = {2016}
}

@book{Gharghabi2020,
author = {Gharghabi, Shaghayegh and Imani, Shima and Bagnall, Anthony and Darvishzadeh, Amirali and Keogh, Eamonn},
booktitle = {Data Min. Knowl. Discov.},
doi = {10.1007/s10618-020-00695-8},
isbn = {1061802000},
issn = {1573756X},
number = {4},
pages = {1104--1135},
publisher = {Springer US},
title = {{An ultra-fast time series distance measure to allow data mining in more complex real-world deployments}},
url = {https://doi.org/10.1007/s10618-020-00695-8},
volume = {34},
year = {2020}
}

@article{Fukunaga1975,
author = {Fukunaga, Keinosuke and Hostetler, Larry D.},
doi = {10.1109/TIT.1975.1055330},
issn = {15579654},
journal = {IEEE Trans. Inf. Theory},
number = {1},
pages = {32--40},
title = {{The Estimation of the Gradient of a Density Function, with Applications in Pattern Recognition}},
volume = {21},
year = {1975}
}

@article{Balasubramanian2013,
author = {Balasubramanian, Arvind},
isbn = {9781450321747},
number = {i},
pages = {1--6},
title = {{Flexible Exploration and Visualization of Motifs in Biomedical Sensor Data}},
url = {papers2://publication/uuid/E3A14BB0-31E2-4439-9133-E5398709323E},
year = {2013}
}

@article{Cole2017,
author = {Cole, Scott R. and Voytek, Bradley},
doi = {10.1016/j.tics.2016.12.008},
issn = {1879307X},
journal = {Trends Cogn. Sci.},
number = {2},
pages = {137--149},
publisher = {Elsevier Ltd},
title = {{Brain Oscillations and the Importance of Waveform Shape}},
volume = {21},
year = {2017}
}

@article{Welford1962,
author = {Welford, B. P.},
doi = {10.1080/00401706.1962.10490022},
issn = {15372723},
journal = {Technometrics},
number = {3},
pages = {419--420},
title = {{Note on a Method for Calculating Corrected Sums of Squares and Products}},
volume = {4},
year = {1962}
}

@article{Rodrigues2021,
author = {Rodrigues, Joao and Probst, Phillip and Gamboa, Hugo},
doi = {10.1109/ICBSII51839.2021.9445154},
isbn = {9781665441261},
journal = {Proc. 2021 IEEE 7th Int. Conf. Bio Signals, Images Instrumentation, ICBSII 2021},
pages = {1--6},
title = {{TSSummarize: A Visual Strategy to Summarize Biosignals}},
year = {2021}
}

@book{Gallier2011,
address = {New York, NY},
author = {Gallier, Jean},
publisher = {Springer New York},
title = {{Discrete Mathematics}},
year = {2011}
}

@article{Sheikh2007,
author = {Sheikh, Yaser and Khan, Erum and Kanade, Takeo},
doi = {10.1109/ICCV.2007.4408978},
journal = {Proc. IEEE Int. Conf. Comput. Vis.},
title = {{Mode-seeking by medoidshifts}},
year = {2007}
}

@article{Parzen1962,
author = {Parzen, E},
journal = {Ann. Math. Stat.},
pages = {1065--1076},
title = {{On the Estimation of Probability Density Functions and Mode}},
volume = {33},
year = {1962}
}

@article{Minnen2007,
author = {Minnen, David and Isbell, Charles L. and Essa, Irfan and Starner, Thad},
isbn = {1577353234},
journal = {Proc. Natl. Conf. Artif. Intell.},
pages = {615--620},
title = {{Discovering multivariate motifs using subsequence density estimation and greedy mixture learning}},
volume = {1},
year = {2007}
}

@article{Zhu2018a,
author = {Zhu, Yan and Yeh, Chin Chia Michael and Zimmerman, Zachary and Kamgar, Kaveh and Keogh, Eamonn},
doi = {10.1109/ICDM.2018.00099},
isbn = {9781538691588},
issn = {15504786},
journal = {Proc. - IEEE Int. Conf. Data Mining, ICDM},
pages = {837--846},
publisher = {IEEE},
title = {{Matrix Profile XI: SCRIMP++: Time Series Motif Discovery at Interactive Speeds}},
year = {2018}
}

@article{Zhu2018,
author = {Zhu, Yan and Zimmerman, Zachary and {Shakibay Senobari}, Nader and Yeh, Chin Chia Michael and Funning, Gareth and Mueen, Abdullah and Brisk, Philip and Keogh, Eamonn},
doi = {10.1007/s10115-017-1138-x},
issn = {02193116},
journal = {Knowl. Inf. Syst.},
number = {1},
pages = {203--236},
publisher = {Springer London},
title = {{Exploiting a novel algorithm and GPUs to break the ten quadrillion pairwise comparisons barrier for time series motifs and joins}},
volume = {54},
year = {2018}
}

@article{Paparrizos2016,
author = {Paparrizos, John and Gravano, Luis},
doi = {10.1145/2949741.2949758},
isbn = {9781450327589},
issn = {01635808},
journal = {SIGMOD Rec.},
number = {1},
pages = {69--76},
title = {{K-Shape: Efficient and Accurate Clustering of Time Series}},
volume = {45},
year = {2016}
}

@article{Bergen2019,
author = {Bergen, Karianne J. and Beroza, Gregory C.},
doi = {10.1007/s00024-018-1995-6},
isbn = {0002401819},
issn = {14209136},
journal = {Pure Appl. Geophys.},
number = {3},
pages = {1037--1059},
title = {{Earthquake Fingerprints: Extracting Waveform Features for Similarity-Based Earthquake Detection}},
volume = {176},
year = {2019}
}

@article{Imani2020,
author = {Imani, Shima and Madrid, Frank and Ding, Wei and Crouter, Scott E. and Keogh, Eamonn},
doi = {10.1007/s10618-020-00702-y},
issn = {1573756X},
journal = {Data Min. Knowl. Discov.},
number = {6},
pages = {1713--1743},
publisher = {Springer},
title = {{Introducing time series snippets: a new primitive for summarizing long time series}},
volume = {34},
year = {2020}
}

@inproceedings{Mendoza-Cardenas2021,
archivePrefix = {arXiv},
arxivId = {2108.03177},
author = {Mendoza-Cardenas, Carlos H. and Brockmeier, Austin J.},
booktitle = {43rd Annu. Int. Conf. IEEE Eng. Med. Biol. Soc.},
eprint = {2108.03177},
month = {aug},
pages = {4},
title = {{Shift-invariant waveform learning on epileptic ECoG}},
url = {http://arxiv.org/abs/2108.03177},
year = {2021}
}

@article{Vedaldi2008,
author = {Vedaldi, Andrea and Soatto, Stefano},
journal = {Lect. Notes Comput. Sci.},
pages = {705--718},
publisher = {Springer Verlag},
title = {{Quick shift and kernel methods for mode seeking}},
volume = {5305},
year = {2008}
}

@article{Aghbari2020,
author = {Aghbari, Zaher Al and Al-Hamadi, Ayoub},
doi = {10.1016/j.procs.2020.03.131},
issn = {18770509},
journal = {Procedia Comput. Sci.},
pages = {595--601},
publisher = {Elsevier B.V.},
title = {{Finding K Most Significant Motifs in Big Time Series Data}},
url = {https://doi.org/10.1016/j.procs.2020.03.131},
volume = {170},
year = {2020}
}

@article{Law2019,
author = {Law, Sean},
doi = {10.21105/joss.01504},
issn = {2475-9066},
journal = {J. Open Source Softw.},
number = {39},
pages = {1504},
title = {{STUMPY: A Powerful and Scalable Python Library for Time Series Data Mining}},
volume = {4},
year = {2019}
}

@article{Westmoreland1996,
author = {Westmoreland, Barbara F.},
doi = {10.4065/71.5.501},
issn = {00256196},
journal = {Mayo Clin. Proc.},
number = {5},
pages = {501--511},
pmid = {8628033},
publisher = {Mayo Foundation for Medical Education and Research},
title = {{Epileptiform electroencephalographic patterns}},
volume = {71},
year = {1996}
}
}

%

\appendix
\section{Detailed QSMP algorithms and GPU parallelism}
\label{sec:supp_algorithms}

\begin{figure}[htb]
    \centering
    \includegraphics[width=0.85\linewidth]{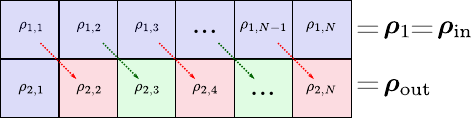}
    \caption{Toy example of the rolling inner-product update
    (\ref{eq:recursive-dot-product-compute}), computing $\rho_{i,j}$ for $i=2$
    and $j=2,\ldots,N$ with two GPU threads (pale red and green). The vector
    $\boldsymbol{\rho}_1$ is precomputed; the buffers
    $\boldsymbol{\rho_\text{in}}$ and $\boldsymbol{\rho_\text{out}}$ hold the
    inner products from the $(i{-}1)$th and $i$th iterations, which enables
    parallelization and GPU acceleration.}
    \label{fig:dot-product-compute}
\end{figure}

For completeness we give the full procedures behind the complexity claims in the
main text. Recall that the squared z-normalized distance
$D_{i,j}$~(\ref{eq:z-normalized_distance}) depends on the subsequences only
through the inner product $\rho_{i,j}=\mathbf{x}_i^\text{T}\mathbf{x}_j$, and
that each $\rho_{i,j}$ follows from $\rho_{i-1,j-1}$ in $\mathcal{O}(1)$
time~(\ref{eq:recursive-dot-product-compute}). The whole $i$th row
$\boldsymbol{\rho}_i=[\rho_{i,1},\ldots,\rho_{i,N}]$ is thus obtained from
$\boldsymbol{\rho}_{i-1}$ in $\mathcal{O}(N)$ time, and this step is
embarrassingly parallel: two buffers $\boldsymbol{\rho}_\text{in}$ and
$\boldsymbol{\rho}_\text{out}$ hold rows $i-1$ and $i$, and the $N$ output cells
are split across GPU threads (Fig.~\ref{fig:dot-product-compute}).

Algorithm~\ref{alg:density} assembles the density estimate
(\ref{eq:density_estimate}) in a single $\mathcal{O}(N^2)$ pass over these
rolling inner products, where \texttt{compute\-Distance} evaluates $D_{i,j}$ from
$\rho_{i,j}$ and the rolling means and standard deviations, and
\texttt{rolling\-Dot\-Product} precomputes $\boldsymbol{\rho}_1$ by FFT. Given the
density, Algorithm~\ref{alg:dist_index_SI} computes the shift-invariant
NN-distance $\mathbf{d}$ and NN-index $\mathbf{v}$, applying the length-$B$
min-pooling filter \texttt{minFilt} to each distance row before the
higher-density nearest-neighbor search, for a total of $\mathcal{O}(N^2(B+3))$
time and $\mathcal{O}(N)$ space. Both algorithms are embarrassingly parallel
(the outer loop across GPUs, the inner loop across the threads of one GPU) and
extend at negligible cost to a grid of $n_\sigma$ kernel widths---the kernel
term $\exp(-D^2/2\sigma^2)$ is $\mathcal{O}(1)$ per $\sigma$---which is what
makes the unsupervised $\sigma$ sweep of the main text nearly free.

\begin{algorithm}[p]
    \scriptsize
    \caption{Computing the density estimate}
    \label{alg:density}
    \DontPrintSemicolon
    \LinesNumbered
    \SetKwComment{comment}{\# }{}
    \SetKwFunction{MeanStd}{rollingMeanAndStd}
    \SetKwFunction{Dot}{rollingDotProduct}
    \SetKwFunction{Zeros}{zeros}
    \SetKwFunction{Distance}{computeDistance}
    \SetKw{To}{$\leftarrow$}
    \SetKwData{rhofirst}{$\boldsymbol{\rho}_1$}
    \SetKwData{rhoin}{$\boldsymbol{\rho_\text{in}}$}
    \SetKwData{rhoout}{$\boldsymbol{\rho_\text{out}}$}
    \SetKwData{mean}{$\boldsymbol{\mu}$}\SetKwData{std}{$\boldsymbol{\sigma}$}
    \SetKwData{fhat}{$\hat{\mathbf{f}}$}
    \KwIn{Time series $\mathbf{x} \in \mathbb{R}^n$, subsequence length $m$,
        and kernel width $\sigma$}
    \KwOut{\fhat, the density estimate}
    $N$ \To $n - m + 1$\;
    \mean, \std $\leftarrow$ \MeanStd{$\mathbf{x}$, $m$}\;
    \rhofirst \To \Dot{$\mathbf{x}[1:m]$, $\mathbf{x}$}\;
    \rhoin \To \rhofirst\;
    \rhoout \To \rhofirst\;
    \fhat \To \Zeros{$N$}\;
    \For{$j\leftarrow 1$ \KwTo $N$}{
        $D$ \To \Distance{$\rhoout[j]$, $\mean[1]$, $\mean[j]$,
            $\std[1]$, $\std[j]$}\;
        $\fhat[j]$ \To $\fhat[j]$ + $\exp(-D^2/{2\sigma^2})$
    }
    \For{$i\leftarrow 2$ \KwTo $N$}{
        $\rhoout[1]$ \To $\rhofirst[i]$\;
        $D$ \To \Distance{$\rhoout[1]$, $\mean[i]$, $\mean[1]$,
            $\std[i]$, $\std[1]$}\;
        $\fhat[1]$ \To $\fhat[1]$ + $\exp(-D^2/{2\sigma^2})$\;
        \For{$j\leftarrow 2$ \KwTo $N$}{
            $\rhoout[j]$ \To $\rhoin[j-1] + \mathbf{x}[i+m-1]\mathbf{x}[j+m-1]
            - \mathbf{x}[i-1]\mathbf{x}[j-1]$\;
            $D$ \To \Distance{$\rhoout[j]$, $\mean[i]$, $\mean[j]$,
                $\std[i]$, $\std[j]$}\;
            $\fhat[j]$ \To $\fhat[j]$ + $\exp(-D^2/{2\sigma^2})$\;
        }
        \rhoin \To \rhoout\;
    }
\end{algorithm}

\begin{algorithm}[p]
    \scriptsize
    \caption{Computing shift-invariant NN-distance and NN-index}
    \label{alg:dist_index_SI}
    \DontPrintSemicolon
    \LinesNumbered
    \SetKwComment{comment}{\# }{}
    \SetKwFunction{MeanStd}{rollingMeanAndStd}
    \SetKwFunction{Dot}{rollingDotProduct}
    \SetKwFunction{Fill}{fill}
    \SetKwFunction{Distance}{computeDistance}
    \SetKwFunction{MinFilt}{minFilt}
    \SetKw{To}{$\leftarrow$}
    \SetKw{And}{\text{and}}
    \SetKwData{rhofirst}{$\boldsymbol{\rho}_1$}
    \SetKwData{rhoin}{$\boldsymbol{\rho_\text{in}}$}
    \SetKwData{rhoout}{$\boldsymbol{\rho_\text{out}}$}
    \SetKwData{mean}{$\boldsymbol{\mu}$}\SetKwData{std}{$\boldsymbol{\sigma}$}
    \SetKwData{fhat}{$\hat{\mathbf{f}}$}
    \SetKwData{NNdist}{$\mathbf{d}$}
    \SetKwData{NNindex}{$\mathbf{v}$}
    \SetKwData{D}{$\mathbf{D}$}
    \SetKwData{DSI}{$\mathbf{D}_{\text{SI}}$}
    \SetKwData{Inf}{\text{Inf}}
    \KwIn{Time series $\mathbf{x} \in \mathbb{R}^n$, subsequence length $m$,
        density \fhat, and min-pooling filter length $B$}
    \KwOut{The NN-distance (\NNdist) and the NN-index (\NNindex)}
    $N$ \To $n - m + 1$\;
    \mean, \std $\leftarrow$ \MeanStd{$\mathbf{x}$, $m$}\;
    \rhofirst \To \Dot{$\mathbf{x}[1:m]$, $\mathbf{x}$}\;
    \rhoin \To \rhofirst\;
    \rhoout \To \rhofirst\;
    \NNdist \To \Fill{N, \Inf}\;
    \NNindex \To \Fill{N, $-1$}\;
    \For{$j\leftarrow 1$ \KwTo $N$}{
        $\D[j]$ \To \Distance{$\rhoout[j]$, $\mean[1]$, $\mean[j]$,
            $\std[1]$, $\std[j]$}\;
    }
    \DSI \To \MinFilt{\D, $B$}\;
    \For{$j \To 1$ \KwTo $N$}{
        \If{$\fhat[j] > \fhat[1]$ \And $\DSI[j] < \NNdist[1]$}{
            $\NNdist[1] = \DSI[j]$\;
            $\NNindex[1] = j$\;
        }
    }
    \For{$i\leftarrow 2$ \KwTo $N$}{
        $\rhoout[1]$ \To $\rhofirst[i]$\;
        $\D[1]$ \To \Distance{$\rhoout[1]$, $\mean[i]$, $\mean[1]$,
            $\std[i]$, $\std[1]$}\;
        \For{$j\leftarrow 2$ \KwTo $N$}{
            $\rhoout[j]$ \To $\rhoin[j-1] + \mathbf{x}[i+m-1]\mathbf{x}[j+m-1]
            - \mathbf{x}[i-1]\mathbf{x}[j-1]$\;
            $\D[j]$ \To \Distance{$\rhoout[j]$, $\mean[i]$, $\mean[j]$,
                $\std[i]$, $\std[j]$}\;
        }
        \DSI \To \MinFilt{\D, $B$}\;
        \For{$j \To 1$ \KwTo $N$}{
            \If{$\fhat[j] > \fhat[i]$ \And $\DSI[j] < \NNdist[i]$}{
                $\NNdist[i] = \DSI[j]$\;
                $\NNindex[i] = j$\;
            }
        }
        \rhoin \To \rhoout\;
    }
\end{algorithm}

\section{Extended recovery experiments: Wavelets with Poisson process activation}
\label{sec:supp_poisson}

All experiments and figures in this supplement are reproducible with the code
at \url{https://github.com/cniel-ud/qsmp/} using referenced scripts in the 
\texttt{scripts/} directory. 

In the main text the \texttt{power-law} dataset tiles the six Morlet wavelets
edge-to-edge, so each length-$m$ analysis window contains exactly one clean
wavelet. Here we stress-test the three methods on a harder variant in which the
$1{,}000$ wavelet arrivals correspond to a {Poisson} process such that activation times are uniformly distributed and neighboring wavelets frequently overlap and
superimpose (about 63\% of consecutive activations overlap). 
We report the mean~$\pm$~95\% CI over 20 seeds for the performance metrics.

\subsection{Quantitative recovery}
\label{sec:supp_table}

Table~\ref{tab:recovery_poisson} scores each method with three metrics: the
number of distinct frequencies recovered (FreqRec, out of~6), the mean
shift-invariant cosine similarity of the matched prototypes (CosSim), and the
mean peak-frequency error (PeakErr). Prototypes are matched to the ground truth
\emph{with replacement}
(each true prototype scored against its closest prediction), so redundancy is
captured only by FreqRec.

\begin{table}[!b]
    \centering
    \caption{Ground-truth recovery on the \texttt{power-law} dataset
    (Poisson-spaced (overlapping) wavelets; mean $\pm$ 95\% CI over seeds).
    $\uparrow$/$\downarrow$: higher/lower is better. FreqRec is out of 6.}
    \label{tab:recovery_poisson}
    \resizebox{\linewidth}{!}{%
    \begin{tabular}{lccc}
    \toprule
    Method & FreqRec $\uparrow$ & CosSim $\uparrow$ & PeakErr (Hz) $\downarrow$ \\
    \midrule
    QSMP & 4.7$\pm$0.4 & 0.83$\pm$0.04 & 14.3$\pm$6.0 \\
    Snippet-Finder & 4.7$\pm$0.4 & 0.77$\pm$0.05 & 19.0$\pm$7.2 \\
    sikmeans & 3.0$\pm$0.2 & 0.56$\pm$0.03 & 39.9$\pm$6.2 \\
    \bottomrule
    \end{tabular}}
\end{table}

QSMP and Snippet-Finder tie on the number of frequencies recovered (a paired
test gives $p=0.87$), but QSMP is significantly better on morphology fidelity:
CosSim $0.83$ vs.\ $0.77$ (paired $t$-test $p=0.035$, Cohen's $d_z=0.51$).
sikmeans is overwhelmed by the prevalence of the low
frequencies and recovers only three of the six patterns on average.

\subsection{Qualitative recovery and failure modes}
\label{sec:supp_figs}

\begin{figure*}[htb]
    \centering
    \includegraphics[width=0.95\textwidth]{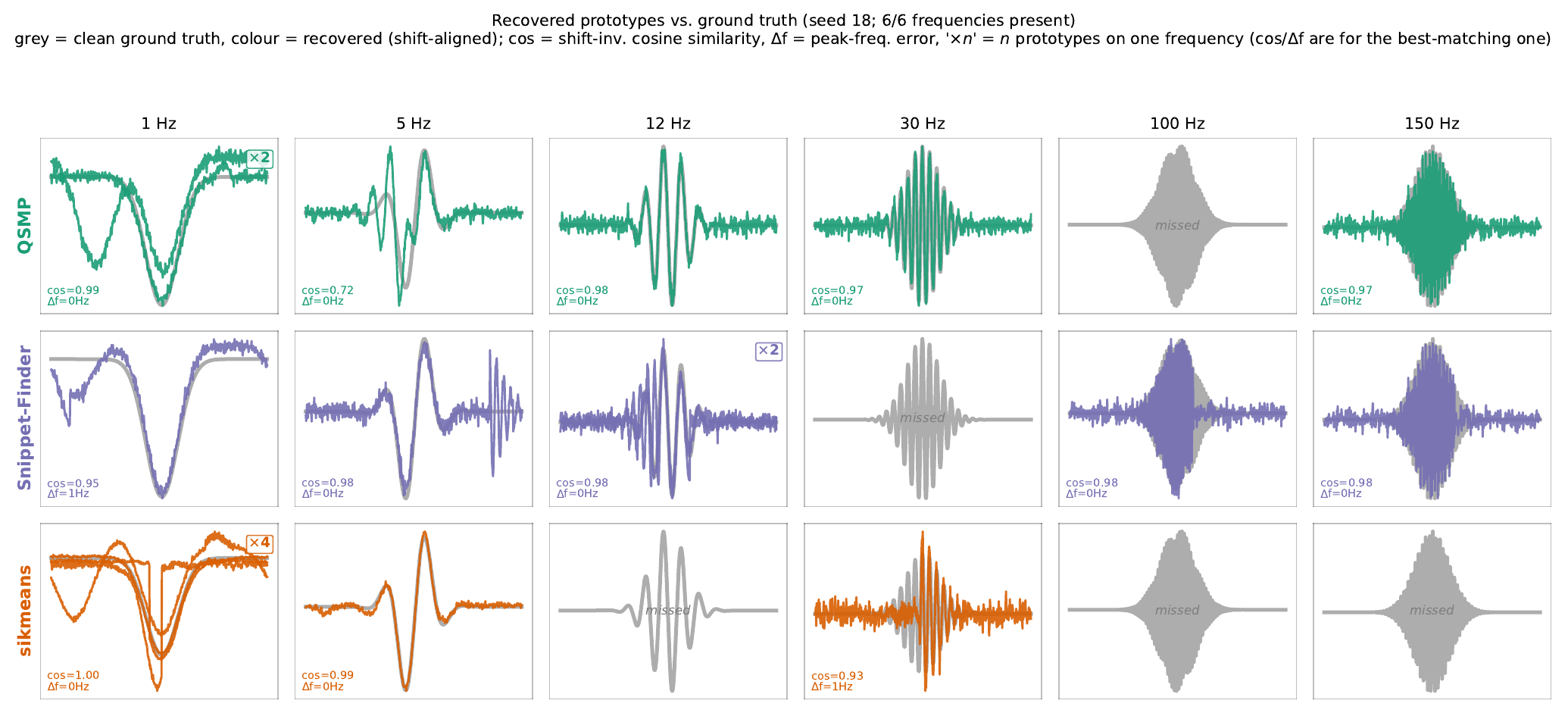}
    \caption{Recovered prototypes vs.\ ground truth for a representative seed.
    One row per method, one column per alphabet frequency; each recovered
    prototype (colour) is shift-aligned and overlaid on the clean ground truth
    (grey). Empty columns are misses; ``$\times n$'' marks $n$ prototypes
    collapsing onto one frequency. Generated by \texttt{viz\_recovery.py}
    (\texttt{gallery\_seed-18}).}
    \label{fig:supp_gallery}
\end{figure*}

The per-seed figures below (Figs.~\ref{fig:supp_gallery},
\ref{fig:supp_averaging}) are drawn on a \emph{representative} seed, chosen
automatically rather than cherry-picked: for each method we take its number of
recovered frequencies (FreqRec) on every seed, and select the single seed whose
per-method FreqRec values are jointly closest to each method's own 20-seed mean
(minimum total absolute deviation across the three methods). The chosen seed is
therefore typical of every method at once, not favourable to any one of them;
for the Poisson set this is seed~18. The aggregate figures and the table use all
20 seeds.

\begin{figure*}[htb]
    \centering
    \hfill\includegraphics[width=0.8\columnwidth]{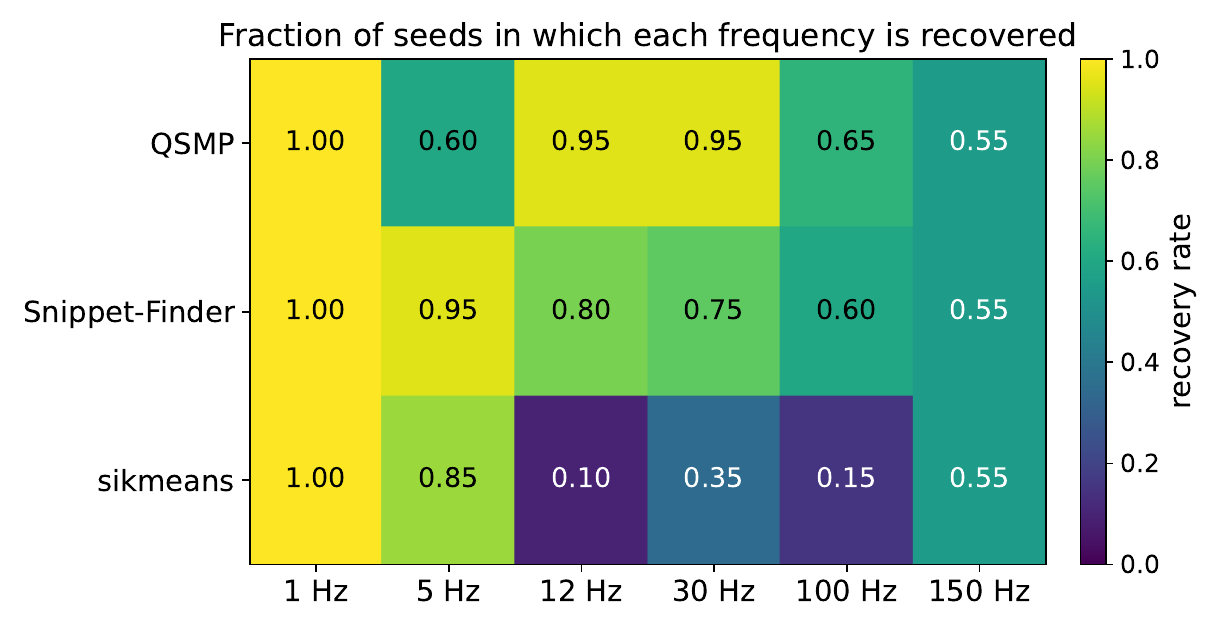}\hfill
    \includegraphics[width=0.8\columnwidth]{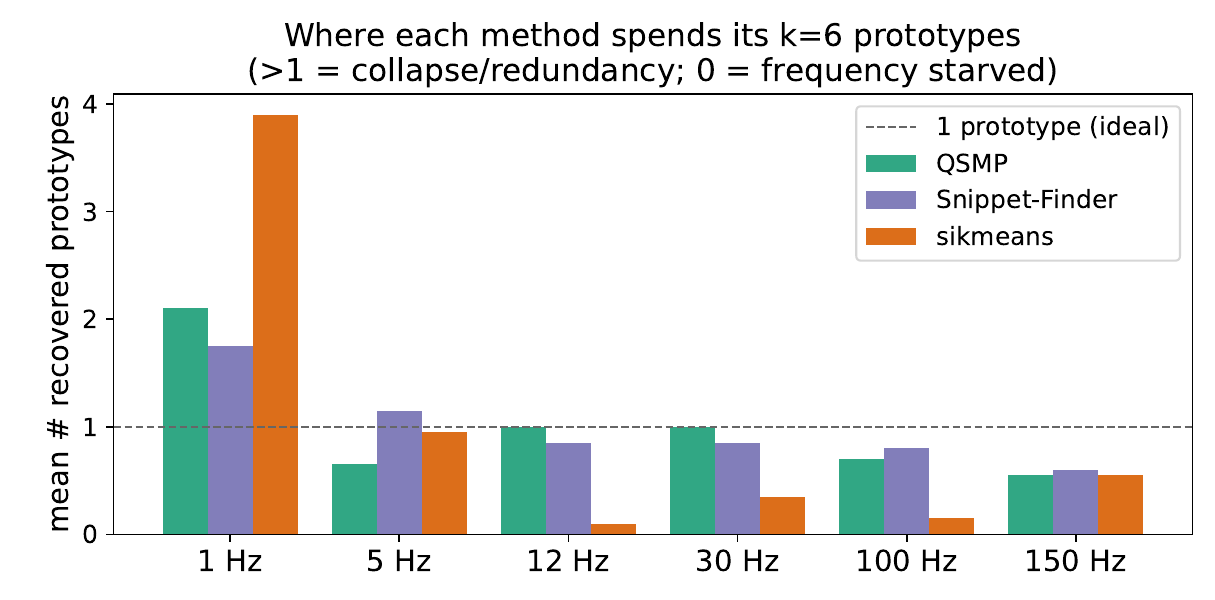}\hfill~\\
    \caption{(Left) Fraction of the 20 seeds in which each method recovers each
    frequency. (Right) Mean number of recovered prototypes snapping to each
    frequency; values above~1 indicate collapse/redundancy, 0 indicates a
    starved frequency. Both from \texttt{viz\_recovery.py}.}
    \label{fig:supp_coverage}
\end{figure*}

\begin{figure*}[htb]
    \centering
    \includegraphics[width=0.95\textwidth]{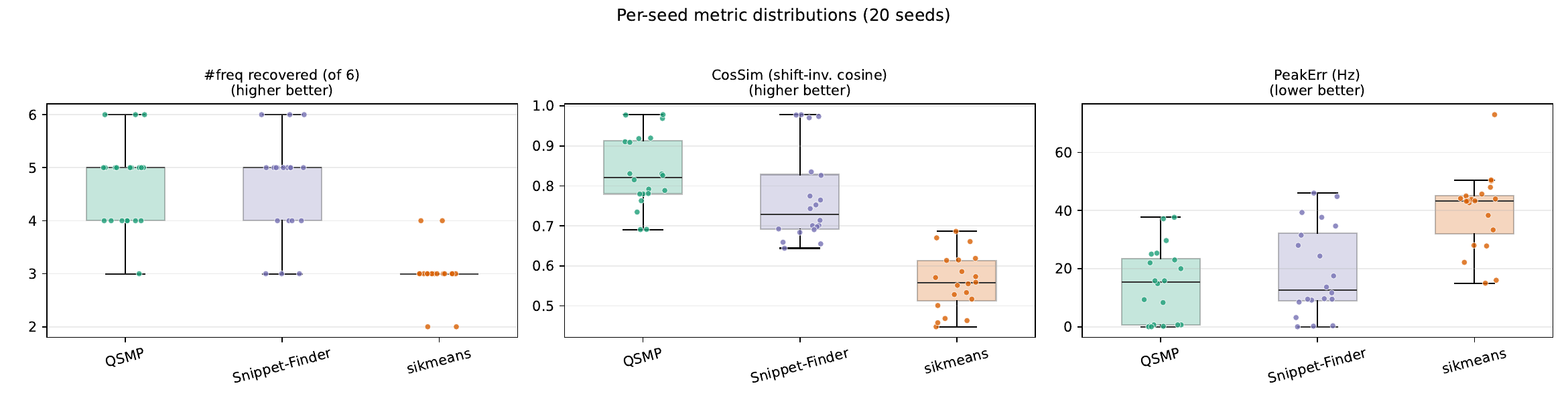}
    \caption{Per-seed spread (strip + box) of FreqRec, CosSim, and PeakErr per
    method over the 20 seeds. From \texttt{viz\_recovery.py}.}
    \label{fig:supp_distributions}
\end{figure*}

Three figures make the table's aggregates concrete. Fig.~\ref{fig:supp_gallery}
shows, on that seed, the prototypes each method returns
shift-aligned and overlaid on the ground truth, one column per frequency: QSMP
and Snippet-Finder both cover most frequencies, whereas sikmeans leaves the rare
high frequencies empty (misses) and stacks several centroids onto the prevalent
low ones ($\times n$ collapse). Fig.~\ref{fig:supp_coverage} aggregates this over
all 20 seeds --- the left panel gives each method's per-frequency recovery rate
(sikmeans rarely reaches 100 or 150~Hz), and the right panel shows where each
method spends its prototypes (sikmeans piles multiple prototypes on 1--5~Hz
while starving the high frequencies, the collapse the main text describes).
Fig.~\ref{fig:supp_distributions} reports the per-seed spread of FreqRec, CosSim
and PeakErr rather than only the mean~$\pm$~CI, confirming the ordering is
consistent across seeds and not driven by outliers.

\subsection{Why Snippet-Finder trails on CosSim}
\label{sec:supp_sf}

Snippet-Finder ties QSMP on the number of frequencies recovered but loses on
CosSim. The reason is structural, and follows from its distance measure. A
Snippet-Finder snippet is a \emph{verbatim} non-overlapping window of the
signal, selected to minimise the MPdist ProfileArea (fidelity + coverage).
MPdist~\cite{Gharghabi2020} considers two windows similar if they share many
similar \emph{sub-subsequences} of length $S=\lceil\text{percentage}\cdot
m\rceil$ (here $S=154$, 30\% of $m$), reporting the 5th percentile of the
fragment-level distances \emph{regardless of order}. Because a match needs only
a small fraction of fragments to agree, MPdist cannot enforce whole-prototype
morphology: a constructed window made of half a 100~Hz wavelet spliced with half
a 12~Hz wavelet scores a \emph{perfect} MPdist match to both prototypes, while
the whole-window cosine correctly penalises it.

Two consequences follow on the Poisson signal. First, among the windows it does
select, Snippet-Finder cannot distinguish a clean prototype window from a
superposition, so it returns contaminated windows: its \emph{median} matched
cosine equals QSMP's ($0.972$), but it has roughly twice the fraction of poor
matches (cosine $<0.5$: 22\% vs.\ 13\%). Second, its coverage objective barely
credits the rarest frequencies (150~Hz occupies $\sim$0.3\% of the signal), so
even its \emph{closest} snippet to 150~Hz is a non-match in 15 of 19 seeds,
sitting at the MPdist noise floor.

As a fairness check we also scored recovery under Snippet-Finder's \emph{own}
distance (mean MPdist to the ground truth, matched with replacement): QSMP still
wins, $3.29$ vs.\ $4.55$ (paired $p=0.003$, $d_z=0.78$). The gap is therefore
not an artifact of evaluating Snippet-Finder with a whole-window measure; it
reflects that Snippet-Finder is coverage oriented rather than a
prototype-recovery method. These numbers are reproduced by
\texttt{sf\_cossim\_analysis.py}.

\subsection{Denoising returned prototypes by shift-aligned averaging}
\label{sec:supp_averaging}

\begin{figure*}[htb]
    \centering
    \includegraphics[width=0.95\textwidth]{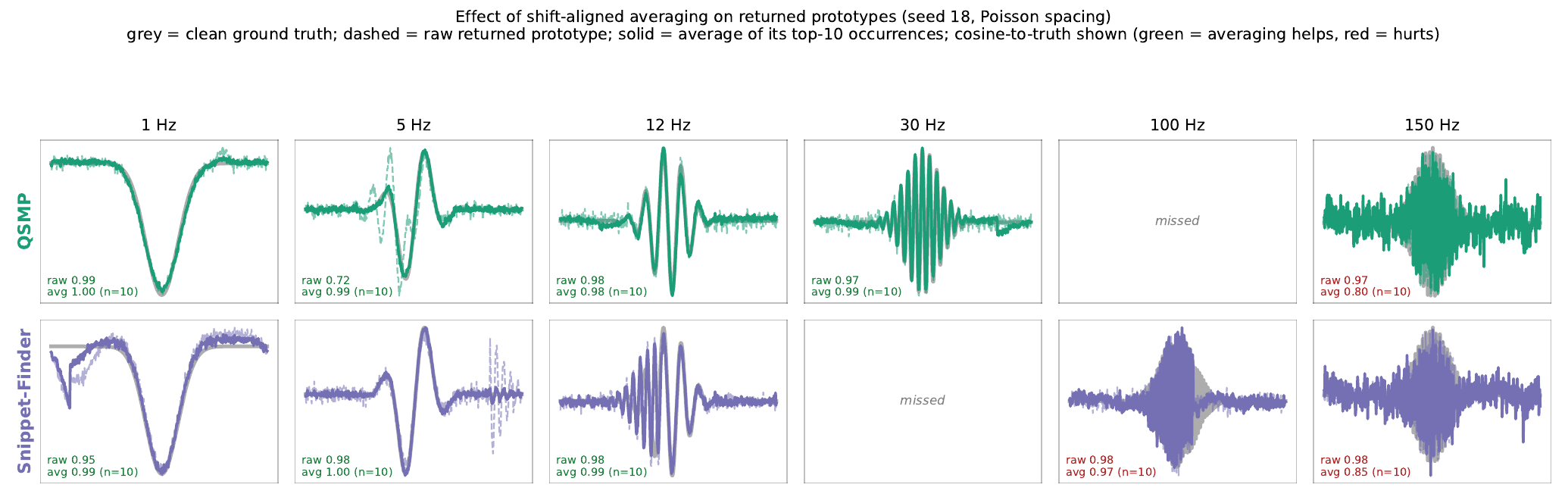}
    \caption{Effect of shift-aligned averaging on the prototypes each method
    returns (representative seed, Poisson spacing). One row per method, one
    column per alphabet frequency. Grey: clean ground truth. Dashed: the raw
    returned prototype (best match for that frequency). Solid: the average of its
    top-$10$ nearest occurrences in the signal. Each cell lists the
    cosine-to-ground-truth of the raw and averaged waveform (green: averaging
    helps; red: it hurts). Averaging denoises the common frequencies but degrades
    the rarest (150~Hz, only 3 activations), where a fixed top-$10$ pulls in
    unrelated windows. Empty cells are frequencies the method missed for this
    seed. Generated by \texttt{scripts/fig\_averaging\_effect.py}.}
    \label{fig:supp_averaging}
\end{figure*}

The raw waveforms both QSMP and Snippet-Finder return are individual (noisy)
subsequences of the signal. A natural question is whether averaging their
occurrences denoises them. We test this directly. For a returned prototype
(a QSMP mode or a Snippet-Finder snippet) we locate its nearest occurrences in
the signal with a sliding-window z-normalised distance profile (the sliding
search supplies the shift; a trivial-match exclusion zone prevents self-overlap),
take a fixed top-$10$, and average the z-normalised windows. The same procedure
is applied to both methods; sikmeans is omitted because its centroids are
already averages. Fig.~\ref{fig:supp_averaging} overlays each raw prototype
(dashed) and its top-$10$ average (solid) on the ground truth (grey).

The effect is two-sided. For the common frequencies, whose occurrences are
plentiful, all ten windows are genuine and averaging cleanly denoises the
waveform (e.g.\ QSMP's 5~Hz mode improves from cosine $0.72$ to $0.99$;
Snippet-Finder's 1~Hz snippet from $0.95$ to $0.99$). A fixed top-$10$
\emph{cannot rescue the rarest} frequencies, however: 150~Hz is activated only 3
times in this seed, so seven of the ten averaged windows are unrelated or
superimposed segments and the average is \emph{worse} than the raw waveform
(QSMP $0.97\!\to\!0.80$, Snippet-Finder $0.98\!\to\!0.85$). This is a limitation
of fixed-count averaging rather than of either discovery method: denoising a
prototype requires enough genuine occurrences to average, which the power-law
prevalence structure denies to the rare high frequencies. An adaptive count
(averaging only occurrences within a similarity threshold) would recover the
common-frequency gains without corrupting the rare ones, at the cost of a
tunable threshold; we report the simple fixed-$10$ variant so the trade-off is
explicit.

\subsection{Limitation of unsupervised \texorpdfstring{$\sigma$}{sigma} selection}
\label{sec:supp_limitation}

QSMP's density modes are individual subsequences of the signal, and the kernel
width $\sigma$ that governs them is selected here without supervision by the
max-min diversity criterion (keep the $\sigma$ whose $k$ modes are most mutually
distinct). However, this metric has a key limitation that we analyzed the results across the 20 seeds
(reproducible via \texttt{sigma\_selection\_analysis.py}). The limitation is that the criterion structurally prefers wide kernels: over the grid
$\sigma\in\{0.5,0.9,1,2,3\}$ it selects only $\sigma=3$ (11 seeds) or $\sigma=2$
(9 seeds) and never explores the small end, because a wider kernel spreads the
modes further apart and mechanically inflates the minimum pairwise distance it
maximizes. Interestingly, the wider $\sigma=3$ recovers slightly
more frequencies than $\sigma=2$ ($4.91$ vs.\ $4.44$ on average), but wider kernel causes redundancy: shape-distinct low-frequency modes collapse onto the same peak, missing a rarer frequency. The limitation can be mitigated by  obtaining the QS-tuple for several $\sigma$ and pick the most
meaningful modes by eye, rather than committing to a single automatically chosen
$\sigma$.

\section{Additional ECoG results (Study019 interictal)}
\label{sec:supp_ecog}

\begin{figure*}[htpb]
    \centering
    \includegraphics[width=0.8\columnwidth]{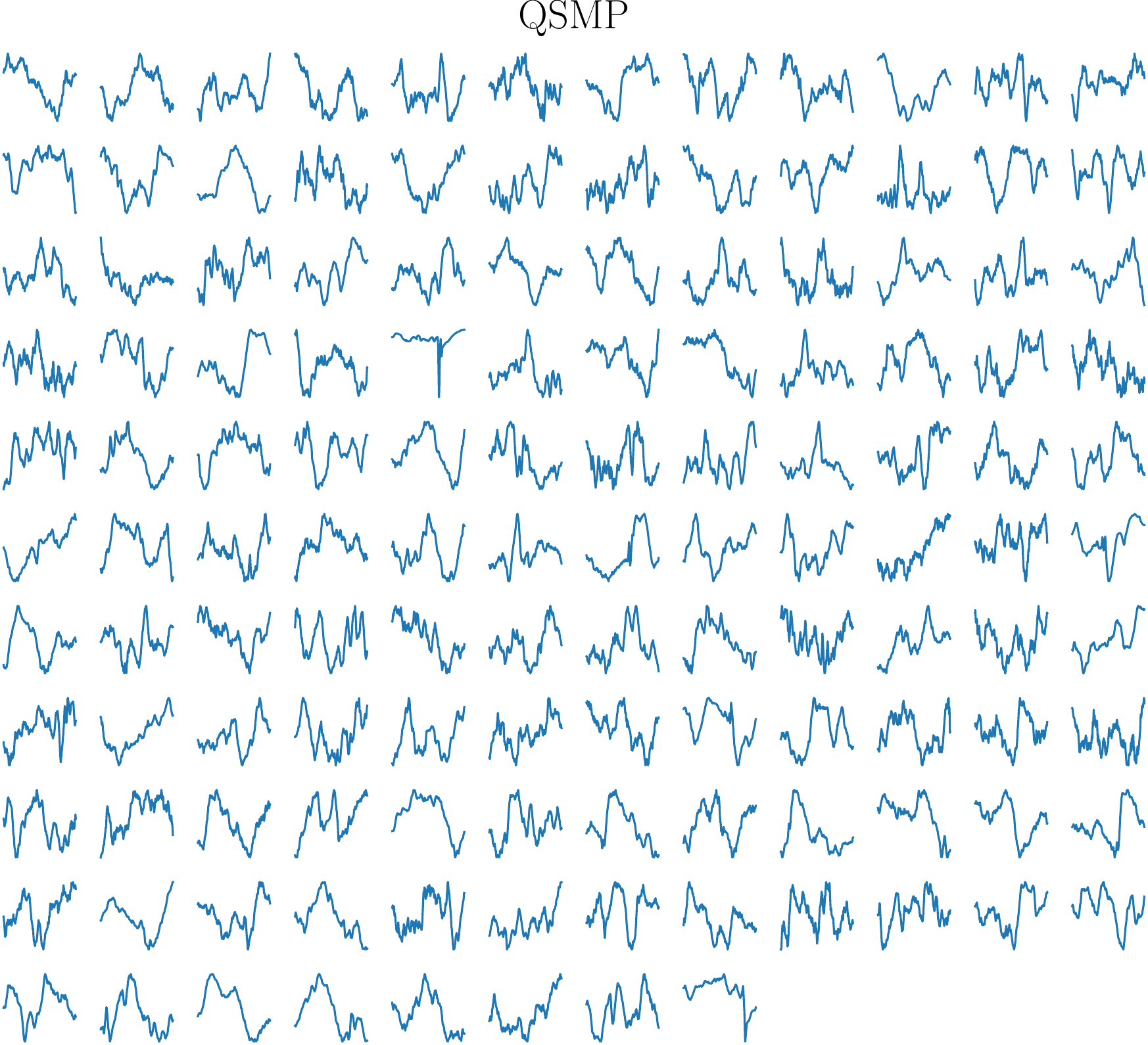}\quad
    \includegraphics[width=0.8\columnwidth]{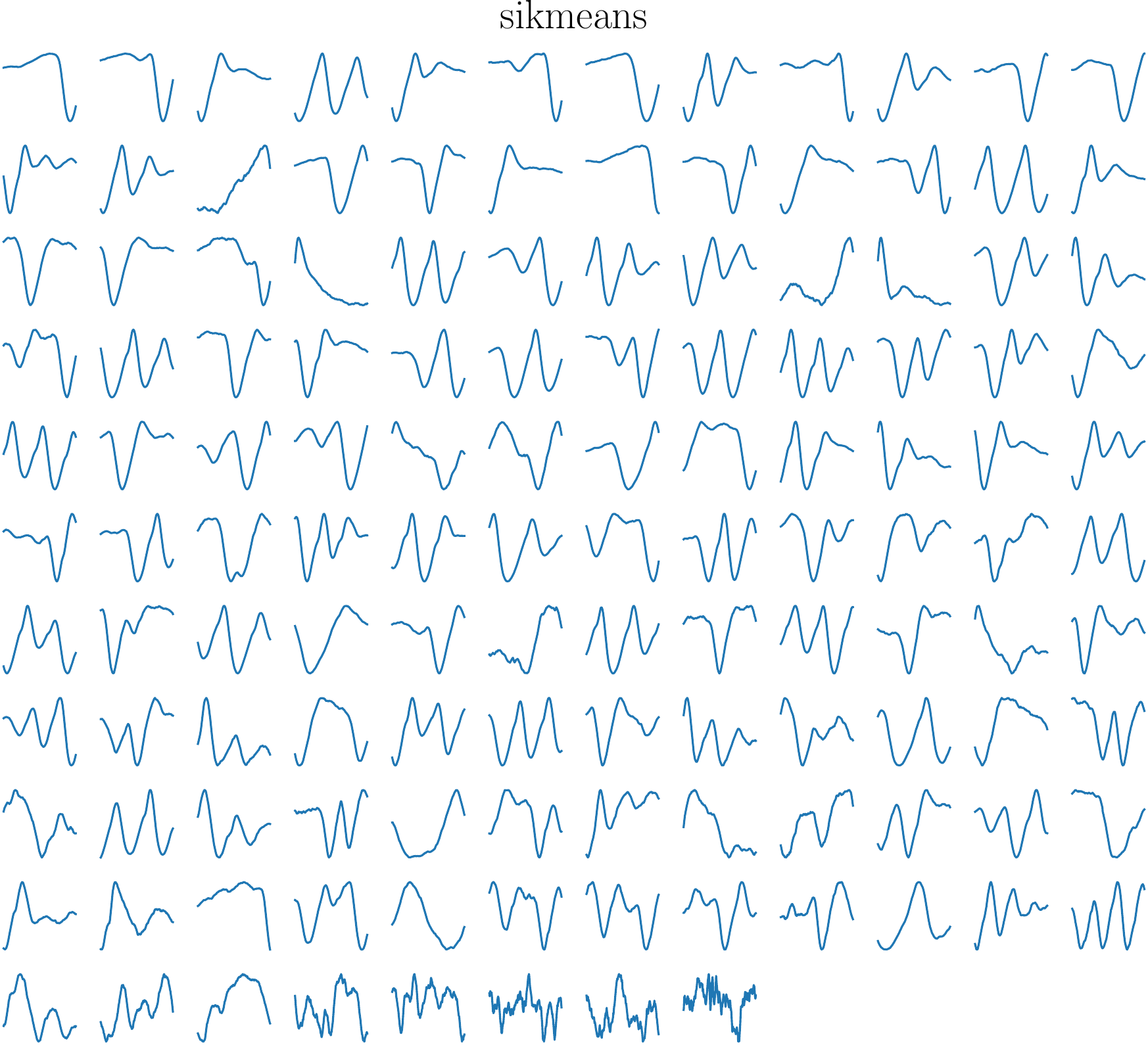}
    \caption{QSMP and sikmeans patterns in the Study019-\emph{interictal}
    dataset.}
    \label{fig:supp_study019-interictal_waves}
\end{figure*}

The main text reports the QSMP vs.\ sikmeans comparison on the \emph{preictal}
Study019 data and notes that the \emph{interictal} data yields the same
qualitative contrast. Fig.~\ref{fig:supp_study019-interictal_waves} shows those
interictal patterns: as in the preictal case, QSMP surfaces sharper,
higher-frequency epileptiform morphologies (its modes are raw subsequences),
whereas sikmeans returns smoother averaged centroids.

\end{document}